\documentclass[journal]{IEEEtran}

\usepackage{graphicx} 
\usepackage{subfigure}
\usepackage{color}
\usepackage{bm}
\usepackage{setspace}
\usepackage[ruled]{algorithm2e}
\usepackage{multirow,subfigure,pifont,hyperref}
\usepackage{verbatim}
\usepackage{amsthm,amsmath,amssymb}
\usepackage{mathrsfs}
\RequirePackage{amsmath}[2016/11/05]

\begin{document}
	\title{Pose-Aided Video-based Person Re-Identification via Recurrent Graph Convolutional Network}
	\author{Honghu Pan, Qiao Liu, Yongyong Chen*, Yunqi He, Yuan Zheng, Feng Zheng, Zhenyu He*,~\IEEEmembership{Senior Member,~IEEE} 
		\thanks{This research was supported by National Natural Science Foundation of China under No. 62172126 and Grant 62106063, by the Shenzhen Research Council under No. JCYJ20210324120202006, by the Special Research project on COVID-19 Prevention and Control of Guangdong Province under No.2020KZDZDX1227, by the Foundation Project of Chongqing Normal University under Grant No. 202109000441, and by the Shenzhen College Stability Support Plan under Grant GXWD20201230155427003-20200824113231001.}
		\thanks{*Corresponding author.}
		\thanks{H. Pan, Y. Chen and  are Z. He with School of Computer Science and Technology, Harbin Institute of Technology, Shenzhen, Shenzhen 518055, China. (Emails: 19B951002@stu.hit.edu.cn, YongyongChen.cn@gmail.com and zhenyuhe@hit.edu.cn)}
		\thanks{Q. Liu is with National Center for Applied Mathematics in Chongqing, Chongqing Normal University, Chongqing 401331, China. (Email: liuqiao.hit@gmail.com)}
		\thanks{Y. He is with College of Information and Computer Engineering, Northeast Forestry University, Harbin 150000, China. (Email: heyunqi.cs@gmail.com)}
		\thanks{Y. Zheng is with School of Computer Science, Inner Mongolia University, Huhehot 010021, China. (Email: zhengyuan@imu.edu.cn)}
		\thanks{F. Zheng is with Southern University of Science and Technology, Shenzhen 518055, China. (Email: zfeng02@gmail.com)}
	}
	
	\maketitle
	
	\begin{figure*}[t]
		\begin{center}
			\includegraphics[width=0.95\textwidth]{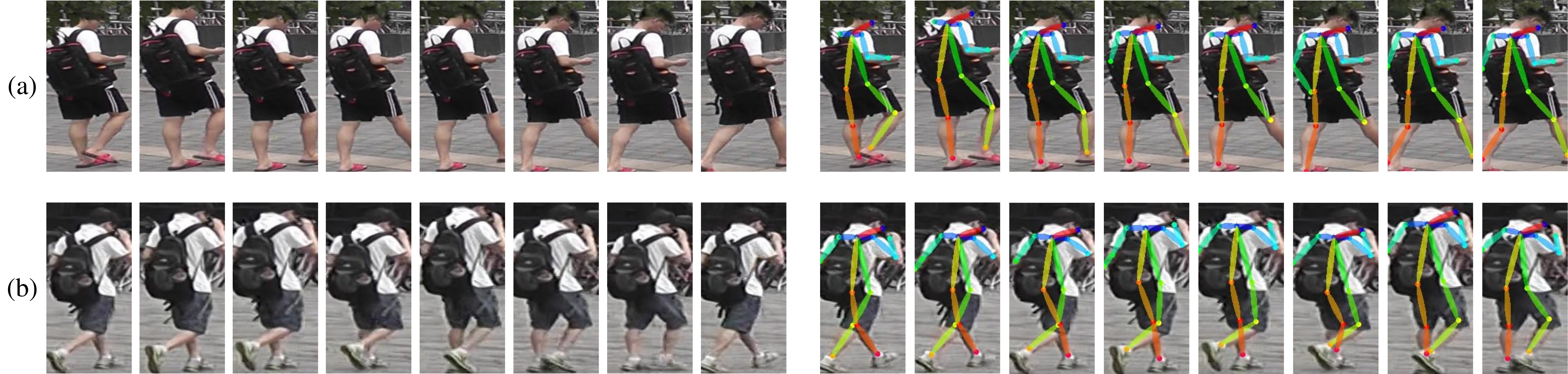}
		\end{center}
		\caption{An illustration of pedestrians with similar appearances.
			In the above figure, the pedestrians of row (a) and row (b) are similar in appearance: they both wear white T-shirts and carry black schoolbags.
			However, they walk with different postures: (a) is playing a mobile phone and (b) is holding his glasses.
			Existing video-based ReID models mainly learn the pedestrian appearance feature, which are difficult to learn a large inter-class variance for pedestrians with similar appearances.
			While in this paper, we propose to learn the pose feature beyond the appearance feature for the pedestrian video retrieval. 
		}
		\label{fig_motivation}
	\end{figure*}

	\begin{abstract}
		Existing methods for video-based person re-identification (ReID) mainly learn the appearance feature of a given pedestrian via a feature extractor and a feature aggregator.
		However, the appearance models would fail when different pedestrians have similar appearances.
		Considering that different pedestrians have different walking postures and body proportions, we propose to learn the discriminative pose feature beyond the appearance feature for video retrieval.
		Specifically, we implement a two-branch architecture to separately learn the appearance feature and pose feature, and then concatenate them together for inference.
		To learn the pose feature, we first detect the pedestrian pose in each frame through an off-the-shelf pose detector, and construct a temporal graph using the pose sequence.
		We then exploit a recurrent graph convolutional network (RGCN) to learn the node embeddings of the temporal pose graph, which devises a global information propagation mechanism to simultaneously achieve the neighborhood aggregation of intra-frame nodes and message passing among inter-frame graphs.
		Finally, we propose a dual-attention method consisting of node-attention and time-attention to obtain the temporal graph representation from the node embeddings, where the self-attention mechanism is employed to learn the importance of each node and each frame.
		We verify the proposed method on three video-based ReID datasets, i.e., Mars, DukeMTMC and iLIDS-VID, whose experimental results demonstrate that the learned pose feature can effectively improve the performance of existing appearance models.
	\end{abstract}

	\section{Introduction}
	\noindent 
	Person re-identification (ReID)~\cite{HMN,SGMN} is an important yet challenging task in public security and video surveillance.
	The visual data provided by the surveillance camera are usually video clips with multiple frames.
	Hence, the video-based person ReID is of great significance and thus has attracted increasing attention in recent years.
	Given a pedestrian video clip with a specific identity, the video-based person ReID aims to find out video clips with the same identity from the retrieval set.
	Existing methods mainly follow the pipeline of ``feature extraction \& feature aggregation''.
	Specifically, they first extract an appearance feature sequence from the input video clip via a deep convolutional neural network (CNN), and then aggregate the feature sequence into a video representation by an aggregator.
	Existing methods have explored multiple aggregators to capture the temporal and spatial clues of pedestrian appearance, such as RNN-based aggregator~\cite{RNN1,RNN2}, attention-based aggregator~\cite{RGSA,STA,ASTA}, 3D convolution-based aggregator~\cite{M3D,AP3D}, and so on.
	
	However, these methods only consider the appearance information of the given pedestrian, which would fail to learn a large inter-class variance for pedestrians with similar appearances as in shown Fig.~\ref{fig_motivation};
	another disadvantage of the appearance feature is the misalignment of human body~\cite{PIE,SSN3D}.
	While the pedestrian pose, which reflects the walking postures and body proportions of pedestrians, could be used as a supplement to alleviate these problems~\cite{PIE,PDC}.
	Thereby, we propose to combine the discriminative pedestrian pose feature with the appearance feature for video-based ReID.
	To this end, we implement a two-branch architecture consisting of an \textbf{appearance feature learning} branch and a \textbf{pose feature learning} one.
	In the training stage, we train each branch separately; while in the inference stage, we concatenate the appearance feature and pose feature as the final video representation for pedestrian retrieval.
	
	The crux is how to learn the pose feature of a given pedestrian.
	We observe that the keypoints of human pose naturally form a graph structure.
	Concretely, nodes in graph refer to the pose keypoints, while edges are defined by the connections between keypoints.	
	Therefore, we can construct a temporal pose graph from the given video clip with multiple frames, while each separate graph in the temporal graph corresponds to a frame.
	The human pose graph constructed in this paper can be regarded as the skeleton graph with 2D node coordinates.
	A large number of studies~\cite{ST-GCN,DGNN,DMGNN} have explored the node embedding learning of temporal skeleton graph for skeleton-based action recognition.
	For example, ST-GCN~\cite{ST-GCN} connects the same joints between consecutive frames and achieves the node embedding learning via a spatial-temporal graph convolutional network.
	However, this node-level connection is not appropriate for pose feature learning, since the pose keypoints in the video frame may be partially occluded due to the camera view and pedestrian posture.
	
	\begin{figure}[t]
		\centering
		\subfigure[Appearance model]{\includegraphics[width=0.23\textwidth]{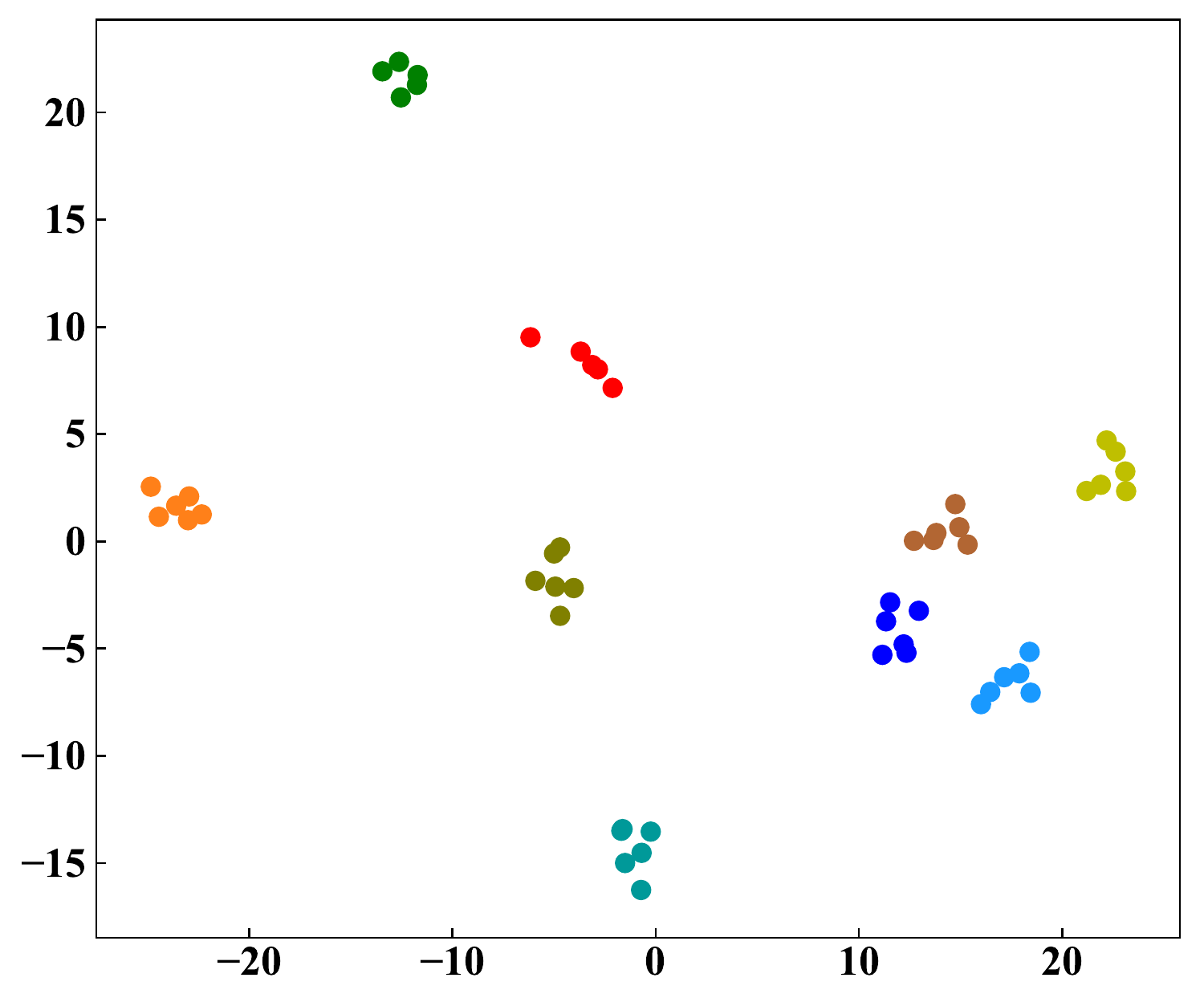}}
		\subfigure[RGCN]{\includegraphics[width=0.23\textwidth]{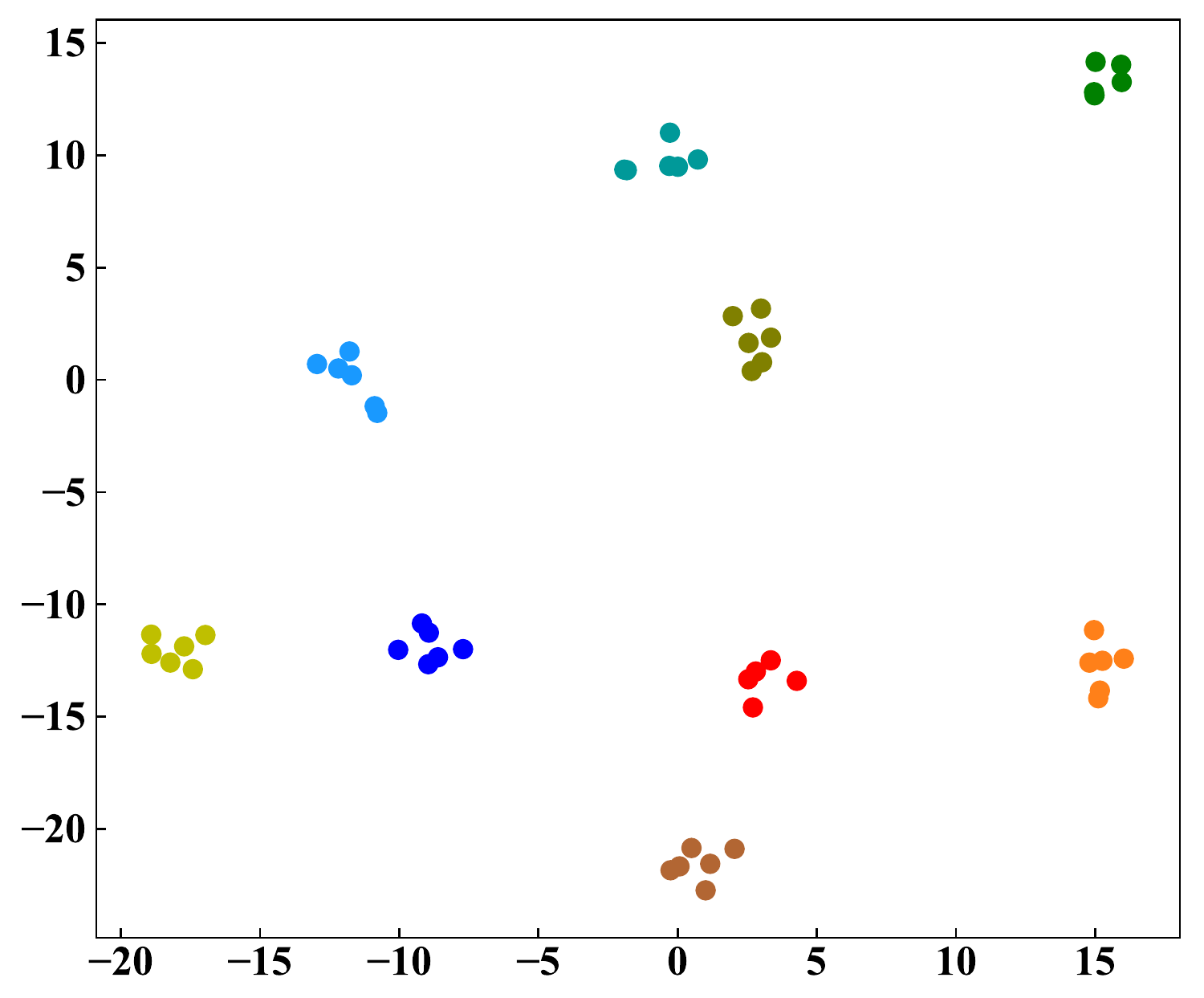}}
		\caption{Visualization of the pedestrian video features learned by (a) existing appearance model and (b) our RGCN.
			We employ t-SNE~\cite{tsne} to perform the data dimensionality reduction on video features, in which marks with the same color belong to the same identity.
			As can be seen, the appearance features mainly distribute in the lower right part of the coordinate axis, while the features involving both appearance and pose information have a larger inter-class variance.
		}
		\label{fig_tsne}
	\end{figure}
	
	Therefore, we propose the \textbf{entire graph message passing} rather than the node-level temporal propagation.
	To this end, we develop a recurrent graph convolutional network (RGCN) to simultaneously perform the node neighborhood aggregation within graph and model the temporal correlations among graphs.
	In RGCN, the hidden state of pose graph is first fed to a  graph convolutional network (GCN), then temporally propagates within a recurrent model.
	Finally, we develop a dual-attention method (DAM), i.e., node-attention and time-attention, to convert the node embeddings into the temporal graph representation or pose feature, in which the node-attention and time-attention take advantage of the self-attention mechanism to evaluate the importance of each node and each frame, respectively.
	
	We note that our RGCN is an add-on method that can be plugged in existing appearance models to remedy the shortcomings of the appearance feature.
	We test our method on the basis of multiple appearance-based baselines, e.g., average pooling, RNN aggregation and attention aggregation.
	The experimental results on three widely-used datasets, i.e., Mars~\cite{Mars}, DukeMTMC~\cite{duke1,duke2} and iLIDS-VID~\cite{ilids}, show that our method outperforms the baseline models by a large margin.
	For example, under the average pooling, the Rank1 on iLIDS-VID increases from 82.0\% to 86.0\% after introducing the pose feature.
	Moreover, the visualization in Fig.~\ref{fig_tsne} demonstrates that the pose information can effectively enlarge the inter-class variance.
	The main contributions of this paper are three-fold:
	\begin{itemize}
		\item To the best of our knowledge, this is the first study that integrates the pedestrian pose feature and the appearance feature simultaneously into one unified framework for video-based ReID;
		\item We propose a RGCN model to learn the node embeddings of the pedestrian pose graph, which combines a recurrent model and GCN to achieve the entire graph message passing mechanism;
		\item We exploit the node-attention and time-attention to obtain the temporal graph representation or pose feature from the node embeddings, where the self-attention mechanism is leveraged to learn the node and frame importance.
	\end{itemize}
	
	The remainder of this paper is organized as follows: 
	Section~\ref{rw} introduces some studies related with this paper; 
	Section~\ref{method} elaborates the two-branch architecture, especially the pose feature learning branch;
	Section~\ref{experiment} presents the experiments and visualizations;
	Section~\ref{conclusion} draws brief conclusions.

	\section{Related Works}
	\label{rw}
	\subsection{Image-based Person Re-identification}
	Image-based ReID aims to identify the identity of a given pedestrian image~\cite{reid_survey,IDM}.
	In this area, learning an image representation~\cite{TCDesc} via a deep convolutional neural network (CNN) is the most widely-used and most successful method.
	The global representation is first applied in image-based ReID.
	For example, TriNet~\cite{TriNet} implemented a hard triplet mining strategy to learn the discirminative features of pedestrian images;
	OSM~\cite{centerlearning} combined the center loss and triplet loss to learn the compact global representations.
	The part-level features have also been explored by existing works:
	PCB~\cite{PCB} first proposed to learn the part-level features through a part-based convolutional module;
	MGH~\cite{MGH} combined the global feature and the part-level features to formulate the multi-granular feature as the pedestrian representation.
	Meanwhile, a great number of studies~\cite{PDC,PNIG,PIE,PVPM} have attempted to learn the pose feature for pedestrian image retrieval:
	PDC~\cite{PDC} used the human body appearance to learn the pose-invariant features;
	PNIG~\cite{PNIG} proposed a pose-normalization GAN~\cite{GAN} model to generate the image via the pedestrian pose;
	PIE~\cite{PIE} employed the pedestrian pose information to reduce the pedestrian misalignment;
	PVPM~\cite{PVPM} proposed a pose-guided visible part matching method for the occluded person ReID.
	Even though the pedestrian pose information has been proven effective in image-based ReID, existing works of video-based ReID have not considered incorporating the pose information into the pedestrian feature.

	\subsection{Video-based Person Re-identification}
	Compared with image-based person ReID, the samples of video-based person ReID contain richer temporal and spatial information.
	Therefore, existing video-based person ReID methods~\cite{RNN1,RNN2,RAFA,FGRA} primarily focus on mining the temporal and spatial correlations via a feature aggregator.
	Early works~\cite{RNN1,RNN2} employed a recurrent model as the aggregator.
	For example, RCN~\cite{RNN1} combined the recurrent neural network and convolutional network to learn the representation of the input video clip;
	RFA-Net~\cite{RNN2} proposed to learn a globally discriminative feature via LSTM~\cite{LSTM}. 
	However, the recurrent model only performs the temporal aggregation on video features, and it fails to capture the spatial clues.
	While the 3D convolution-based aggregator can model both temporal and spatial information: M3D~\cite{M3D} inserted the 3D convolutional layers into the 2D CNN model to enable the multi-scale feature learning; AP3D~\cite{AP3D} implemented an appearance-preserving module by using the 3D convolution; SSN3D~\cite{SSN3D} adopted the 3D convolution across frames to alleviate the temporal appearance misalignment.
	The disadvantage of the 3D convolution is the high space complexity, while the attention-based aggregator~\cite{RGSA,STA,ASTA} can effectively reduce storage: RGSA~\cite{RGSA} implemented a relation-guided spatial attention module and a relation-guided temporal refinement module to exploit both the spatial and temporal clues; STA~\cite{ASTA} enabled the multi-granularity feature aggregation via a spatial-temporal attention approach;
	CPA~\cite{CPA}.
	However, the above-mentioned methods only consider the pedestrian appearance, while this paper proposes to learn both the pose and appearance features for video retrieval.

	\subsection{Graph Convolutional Network}
	The GCN models~\cite{GCN,GCN1,GCN2} usually refer to the spectral-based graph neural networks, which define the node neighborhood aggregation via the graph Laplacian decomposition.
	SpecNet~\cite{GCN1} proposed to learn the graph filter directly;
	to reduce the learning parameters and parameterize the graph filter, ChebNet~\cite{GCN2} approximated the graph filter with the Chebyshev polynomial;
	GCN~\cite{GCN} proposed a layer-wise architecture to learn the graph embedding, where each layer aggregates 1-hop neighborhood for each node.
	Recently, a lot of studies~\cite{AAGCN,STGCN,MGN} successfully applied the GCN model in person ReID.
	In image-based ReID, AAGCN~\cite{AAGCN} took advantage of the low-pass property of GCN to reduce the intra-class variance;
	HLGAT~\cite{HLGAT} proposed a local graph attention network to learn the intra-local and  inter-local relations.
	In video-based ReID,
	CTL~\cite{CTL} employed a graph network to learn the corrections among human parts, where the nodes are extracted via a pose estimator;
	STGCN~\cite{STGCN} implemented a temporal-spatial GCN to aggregate the temporal and spatial features;
	MGN~\cite{MGN} proposed the multi-granular feature aggregation via the hyper-graph model.
	While in this paper, we combine the recurrent models and GCN to learn the pedestrian pose feature for video-based ReID.

	\subsection{Skeleton-based Action Recognition}
	The skeleton-based action recognition methods using deep learning algorithms can be summarized as three categories: CNN-based~\cite{CNN-SK1,CNN-SK2}, RNN-based~\cite{RNN-SK1,RNN-SK2,RNN-SK3} and GCN-based~\cite{ST-GCN} models.
	The CNN-based models transform the 3D skeleton into the image data, then employ a CNN framework to learn the feature.
	For example, MTLN~\cite{CNN-SK1} generated three video clips from each skeleton sequence, where each clip corresponds to one channel of the origin skeleton data;
	ESV~\cite{CNN-SK2} proposed an enhanced skeleton visualization method that converts the skeleton sequence into the 2D images.
	The CNN-based models could not capture the temporal correlations, while the RNN-based ones can model the continuous motion.
	EleAtt~\cite{RNN-SK1} combined the attention mechanism and a recurrent model to learn the importance of each element and model the temporal dynamics;
	SR-TSL~\cite{RNN-SK2} proposed a spatial reasoning module and a spatial reasoning module to achieve the spatial and temporal message passing;
	VA-LSTM~\cite{RNN-SK3} devised a LSTM-based adaptive framework to reduce view
	variations by automatically regulating observation viewpoints.
	Both the CNN-based~\cite{CNN-SK1,CNN-SK2} and RNN-based~\cite{RNN-SK1,RNN-SK2,RNN-SK3} models fail to exploit the skeleton topology, i.e., the joint connections.
	In recent years, the GCN-based models have shown its advantage in skeleton-based action recognition since the graph models can explicitly learn the node corrections from the skeleton data.
	ST-GCN~\cite{ST-GCN} presented the first strong baseline: it constructed a spatial-temporal graph for each skeleton sequence, then learned the node features by spatial and temporal aggregation.
	Many graph-based studies have been proposed on the basis of ST-GCN:
	DGNN~\cite{DGNN} constructed a directed acyclic graph to model the relationship between
	joints and bones;
	DMGNN~\cite{DMGNN} constructed a multiscale graph to learn the part-level relations;
	Shift-GCN~\cite{Shift-GCN} integrated the shift operation and GCN to implement a lightweight architecture.
	The main difference between this task and our pose feature learning lies in the data source:
	the skeleton data~\cite{kinetics} are collected by the depth sensor, while ours are detected by a pose detector.
	Hence, our pose data would be partially occluded.
	This paper devises a global information propagation mechanism to alleviate the influence from local node occlusion.

	\begin{figure*}[t]
		\centering
		\begin{center}
			\includegraphics[width=0.98 \textwidth]{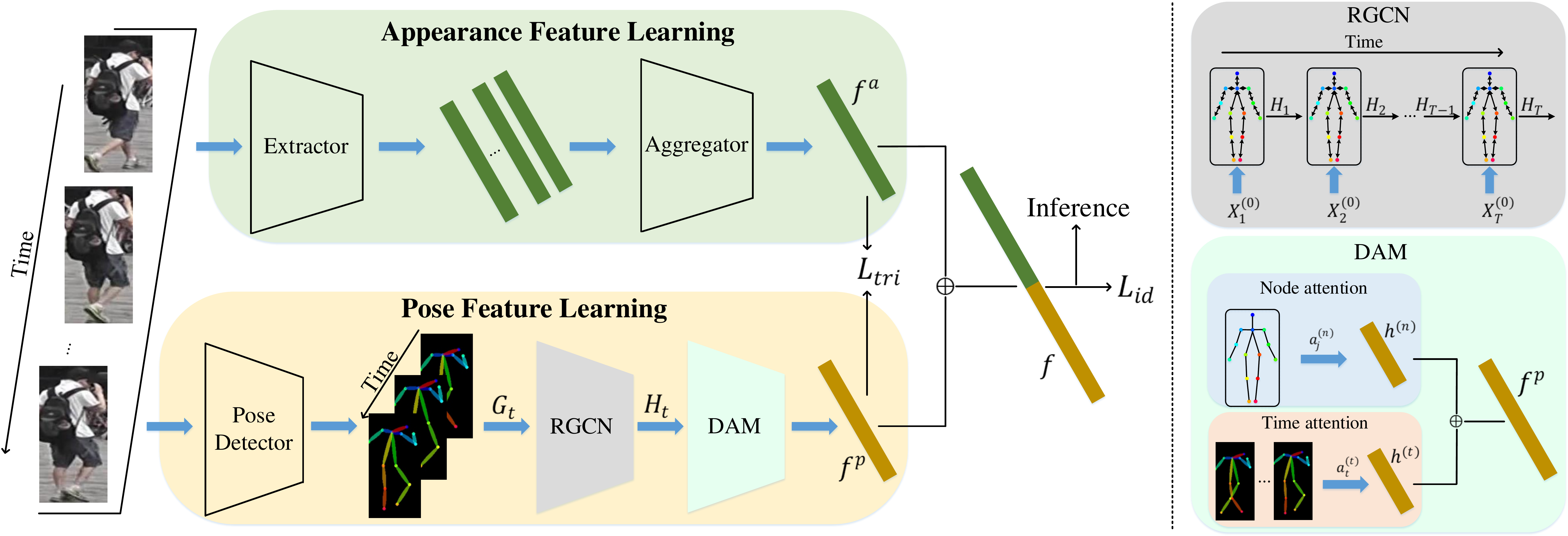}
		\end{center}
		\caption{The architecture of our model consists of an appearance feature learning module and a pose feature learning one, where $\oplus$ denotes the concatenation operation.
			The appearance feature learning module learns the appearance feature $f^a$ via a feature extractor and a feature aggregator.
			The pose feature learning module first constructs a temporal pose graph $G=\{G_t | t=1,2,\cdots,T\}$ via an off-the-shelf pose detector, then employs the proposed RGCN to learn the node embeddings $H=\{H_t\}$, last utilizes the dual-attention methods (DAM) to learn the pose feature $f^p$.
			The RGCN and DAM are shown in the left part of the above figure: RGCN simultaneously aggregates node neighborhood information and propagates graph hidden state $H_t$;	DAM converts the node features into $f^p$ via the node-attention and time-attention.
			In the training stage, we train $f^a$ and $f^p$ separately; while in the inference stage, we concatenate $f^a$ and $f^p$ as the video representation $f$.
		}
		\label{fig_framework}
	\end{figure*}

	\section{Methodologies}
	\label{method}
	In this section, we first present the definition of video-based person ReID and the overall framework of the proposed model in Section~\ref{PDaO}, then introduce the appearance feature learning baselines in Section~\ref{AFL}, next elaborate the pose feature learning paradigm by RGCN in Section~\ref{PFL}, last define the loss functions for model training in Section~\ref{MT}.
	
	\subsection{Problem Definition and Overview}
	\label{PDaO}
	To enable the pedestrian video retrieval, the representation learning is widely employed in video-based ReID.
	Specifically, we learn a representation vector for each video clip by using deep neural networks, whose objective is to maximize the similarity between intra-class samples and minimize that between inter-class samples.
	We denote the given video clip as $V=\{I_1, I_2,\cdots, I_T\}$, where $I_t$ is the $t$-th frame and $T$ is the length.
	We aim to learn a discriminative feature $f$ from the image sequence $V$.
	
	Existing methods mainly focus on the appearance feature of the pedestrian, which follow the pipeline of ``extraction \& aggregation''.
	Generally, a backbone network (e.g., ResNet~\cite{ResNet}) is first adopted to extract the appearance feature $f^a_t$ of the image sequence $I_t$, then an aggregator, e.g., average pooling, RNN-based aggregator or attention-based aggregator, performs the feature aggregation from the feature sequence $f^a_t$ to the video appearance representation $f^a$.
	
	However, existing methods only consider appearances of pedestrians, which would hardly learn a large inter-class variance for pedestrians with similar appearances.
	To resolve this issue, we propose the pose-aided video-based person ReID, in which the pose information is involved in the learned video representation.
	To this end, we propose a two-branch architecture as in shown Fig.~\ref{fig_framework} to separately learn the appearance feature $f^a$ and pose feature $f^p$.
	The main contribution of this paper is the learning paradigm of $f^p$.
	As can be seen in Fig.~\ref{fig_framework}, we first construct a temporal pose graph with the detected pedestrian pose, then propose a recurrent graph convolutional network (RGCN) model to learn the node embeddings of the temporal graph, last exploit the dual-attention (DAM) mechanism to obtain the graph representation.
	
	In this paper, we denote the matrices and vectors as uppercase characters and lowercase characters, respectively.
	The notations and their corresponding descriptions are presented in Table~\ref{table_notation}.
	
	\begin{table}[t]
		\small
		\centering
		\caption{Notations and descriptions.}
		\begin{spacing}{1.25}
			\begin{tabular}{cl}
				\hline
				Notations & Descriptions \\ \hline
				$V$	& The input video            \\  
				$T$ & The length of $V$            \\
				$I_t$ & The $t$-th frame of $V$            \\
				$f^a$ & The appearance feature            \\
				$f^p$ & The pose feature          \\
				$G$ & The temporal pose graph            \\
				$G_t$ & The $t$-th frame pose graph in $G$           \\
				$X_t^{(0)}$ & The initialized feature of $G_t$           \\
				$h_t$ & The hidden state of a sole node           \\
				$H_t$ & The hidden state of $G_t$        \\
				$a_j^{(n)}$ & The node weight learned by node attention        \\ 
				$a_t^{(n)}$ & The frame weight learned by time attention        \\
				$h^{(n)}$ & The graph feature learned by node attention        \\ 
				$h^{(t)}$ & The graph feature learned by time attention        \\ 
				\hline
			\end{tabular}	
		\end{spacing}
		\label{table_notation}
	\end{table}
	
	\subsection{Appearance Feature Learning}
	\label{AFL}
	In this section, we introduce several appearance-based baselines to make this paper self-contained.
	Specifically, we present three widely-used aggregators for the appearance feature learning:
	
	\noindent \textbf{a) Average Pooling (AP)} takes the average value of $f^a_t$ as the final representation $f^a$:
	\begin{equation}
	f^a = \frac{1}{T} \sum_{t=1}^T f^a_t .
	\label{eq_ap}
	\end{equation}
	
	\noindent \textbf{b) Attention Aggregation (AA)} performs the weighted average on $f^a_t$, where the weight coefficients $a_t$ are learned by the attention mechanism ($\sum_{t=1}^T a_t=1$):
	\begin{equation}
	f^a = \sum_{t=1}^T a_t f^a_t .
	\label{eq_aa}
	\end{equation}
	
	\noindent \textbf{c) RNN Aggregation (RA)} utilizes a recurrent model (i.e., LSTM~\cite{LSTM}) to capture the temporal clues:
	\begin{equation}
	f^a = \frac{1}{T} \sum_{t=1}^T o^a_t ,
	\label{eq_ra}
	\end{equation}
	where $o^a_t$ denotes the output of the recurrent model in the $t$-th frame.
	
	\subsection{Pose Feature Learning}
	\label{PFL}
	In this section, we first introduce the construction of the temporal pose graph, then elaborate the RGCN model for node embeddings learning, last present the dual-attention method for graph representation learning.
	
	\subsubsection{Graph Construction}
\begin{figure}[!t]
		\centering
		\subfigure[Pose graph]{\includegraphics[width=0.13\textwidth]{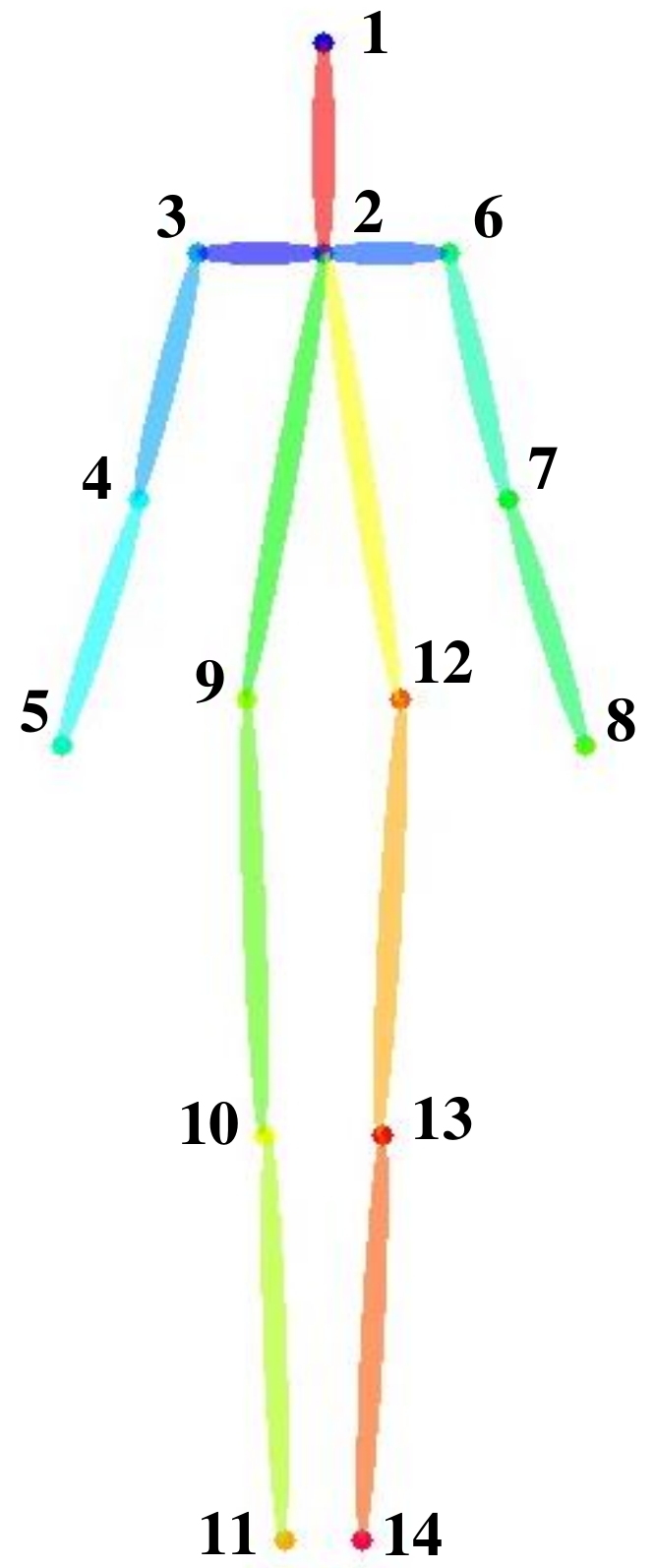}
			\label{fig_graph_a}}
		\subfigure[Adjacency matrix]{\includegraphics[width=0.25\textwidth]{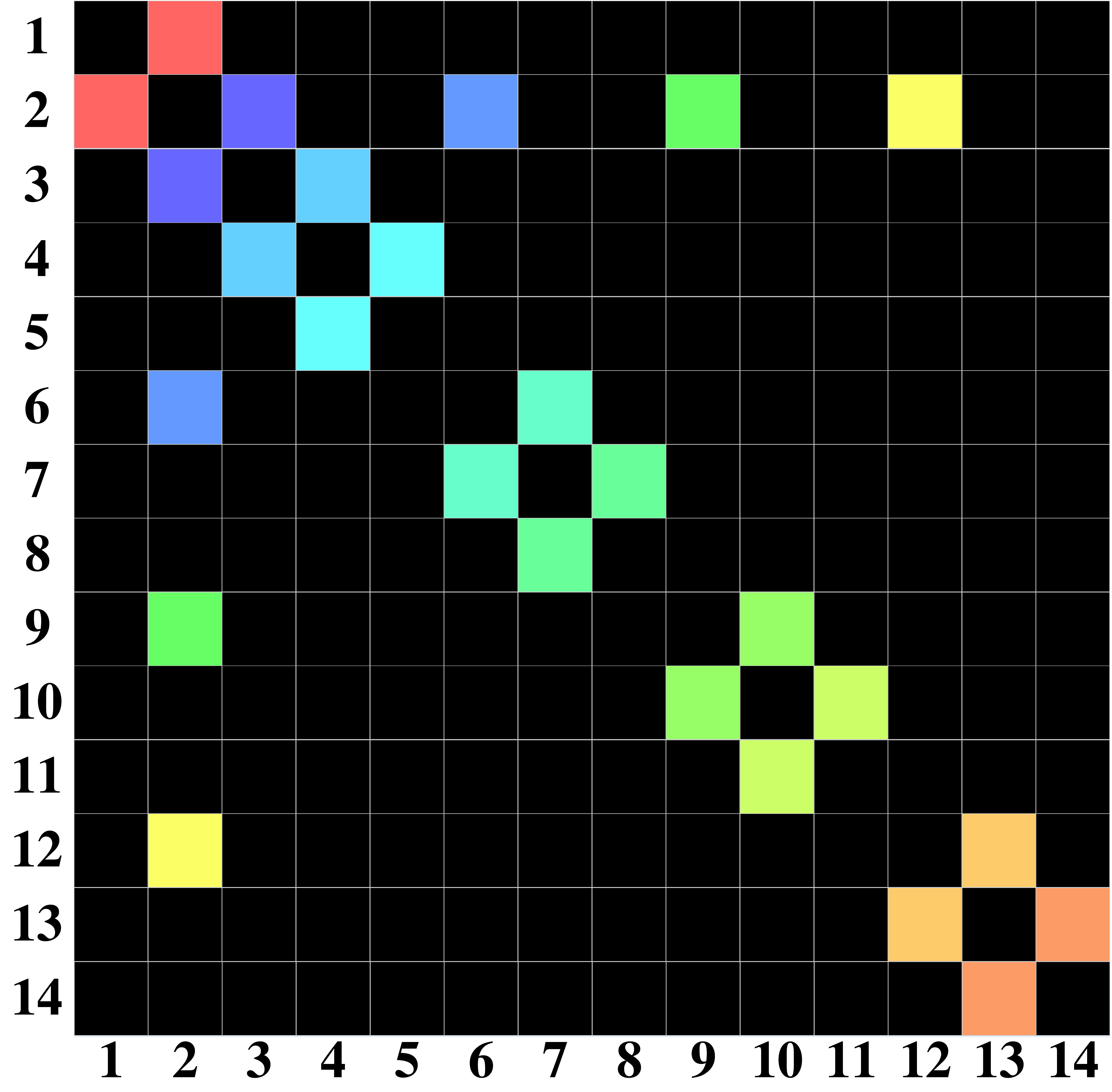}
			\label{fig_graph_b}}
		\caption{The visualization of the pose graph and its adjacency matrix. 
			The pose graph contains 14 nodes, and each of them refers to a pose keypoint on the human body.
			The adjacency matrix consists of binary elements, in which the colored grid and black grid denote 1 and 0, respectively.
		}
		\label{fig_graph}
	\end{figure}

	Given a video clip containing $T$ frames, we employ OpenPose~\cite{OpenPose}, a popular 2D pose estimator, to detect the pedestrian pose in each frame.
	As seen in Fig.~\ref{fig_graph_a}, we retain 14 keypoints for each pedestrian, while each keypoint represents a specific part; for instance, keypoint 1 refers to the head, 3 and 6 refer to the shoulder.
	We then establish the connection between the keypoints according to the connections of human skeleton, for example, 1 and 2, 2 and 3, and etc.
	
	With the pose keypoints and the connections, we construct a pose graph $G_t$ in the $t$-th frame, where $G_t$ is composed of the nodes $V_t$ and the edges $E$.
	Nodes $V_t$ consists of the detected keypoints, i.e., $V_t=\{v_{ti} | i=1,2,\cdots,14\}$.
	The feature of $v_{ti}$ is initialized to the 2D position of keypoint $v_i$ in the $t$-th frame.
	For keypoints that are not detected due to the occlusion, we use zero padding to define their initial features.
	Edges $E$ depict the connections between pairwise nodes.
	We employ an adjacency matrix $A=\{A_{ij}\}$ to represent $E$, where $A_{ij}$ is equal to 1 if $v_{ti}$ is connected to $v_{tj}$ and 0 if not.
	As can be seen in Fig.~\ref{fig_graph_b}, the adjacency matrix is a symmetrical binary matrix, and it is constant over time.
	For a video clip, we could obtain $T$ pose graphs and then combine them together to form a temporal pose graph $G=\{G_t | t=1,2,\cdots,T\}$.

	\subsubsection{Recurrent Graph Convolutional Network}
	To learn the node embeddings of the continuous pose graph $G=\{G_t | t=1,2,\cdots,T\}$, we should consider both the node features within $G_t$ and the temporal information among $G_t$.
	A simple idea is to follow the GCN\&RNN pipeline, which first learns the $t$-th frame node embeddings with a GCN model:
	\begin{equation}
	X_t = \sigma (\tilde{D}^{-1/2} \tilde{A} \tilde{D}^{-1/2} X_t^{(0)} W_g) ,
	\label{eq_GCN}
	\end{equation}
	and then employs a recurrent model to pass the hidden state of each node over time:
	\begin{equation}
	h_t = {\rm tanh} (W_h h_{t-1} + W_x x_{it} + b).
	\label{eq_RNN}
	\end{equation}
	In Eq.~(\ref{eq_GCN}), $X_t^{(0)} \in R^{14 \times 2}$ denotes the initialized node features in the $t$-th frame; $X_t = \{x_{it} | i=1,2,\cdots,14\}$ denotes the graph embedding after GCN model; $\sigma(\cdot)$ refers to the {\rm ReLU} activation function; $\tilde{A}=A+I_n$; $\tilde{D}_{ii} = \sum_{j} \tilde{A}_{ij}$; \textcolor{blue}{ $I_n$ is an identity matrix with size $n \times n$;} and $W_g \in R^{2 \times m} $ is the learnable parameter.
	In Eq.~(\ref{eq_RNN}), $x_{it} \in R^{m \times 1} $ denotes the $i$-th node feature in the $t$-th frame; $h_t \in R^{n \times 1}$ refers to the hidden state; ${\rm tanh}$ is the activation function; $W_h \in R^{n \times n} $, $W_x \in R^{n \times m} $ and $b$ are the learnable parameters.
	
	However, this GCN\&RNN pipeline requires us to input the nodes one by one into the RNN model, which is very time-consuming.
	Moreover, this pipeline performs the node message passing like ST-GCN~\cite{ST-GCN}.
	When the occlusion occurs, it fails to capture the accurate node feature.
	To reduce the time complexity and alleviate the influence of the keypoint occlusion, we propose the recurrent graph convolution networks (RGCN), whose recurrent model takes the entire graph feature rather than a sole node feature as input.
	In the RGCN, the GCN model and the RNN model share the learnable weight $W_x$:
	\begin{equation}
	H_t = {\rm tanh} (H_{t-1} W_h + \sigma (\hat{A} X_t^{(0)} W_x) + b).
	\label{eq_RGCN}
	\end{equation}
	We denote $\tilde{D}^{-1/2} \tilde{A} \tilde{D}^{-1/2}$ as $\hat{A}$ for short.
	In Eq.~(\ref{eq_RGCN}), $H_t \in R^{14 \times n}$ denotes the hidden state of the entire graph in the $t$-th frame; $W_h \in R^{n \times n} $, $W_x \in R^{2 \times n} $ and $b$ are the learnable parameters.
	We note that Eq.~(\ref{eq_RGCN}) can also be extended to the multi-layer graph convolution (GC) structure. For example, RGCN with two GC layers can be denoted as:
	\begin{equation}
	H_t = {\rm tanh}(H_{t-1} W_h + \sigma (\hat{A}\hat{A} X_t^{(0)} W_x^{(1)} W_x^{(2)}) + b),
	\label{eq_RGCN3}
	\end{equation}
	where $W_x^{(1)}$ and $W_x^{(2)}$ denote the learnable parameters in the first and second layer, respectively.
	RGCN model with more than two GC layers could also be implemented by the above discipline.
	
	\textbf{Discussion.}
	Compared to the GCN\&RNN pipeline and ST-GCN~\cite{ST-GCN}, our RGCN has the following advantages:
	1) the hidden state $H_t$ of RGCN in Eq.~(\ref{eq_RGCN}) contains the global information of the pose graph; 
	2) RGCN runs faster,as it avoids to input the node one-by-one into model; 
	3) RGCN contains fewer parameters, which benefits from the weight-sharing.
	
	RGCN of Eq.~(\ref{eq_RGCN}) combines GCN and the vanilla RNN model, while the vanilla RNN can also be replaced by other recurrent models, such as LSTM~\cite{LSTM}.
	In this paper, we also implement an enhanced model using LSTM and GCN, whose equations are presented as follows:
	\begin{equation}
	\begin{split}
	& F_t = {\rm sigmoid}(\sigma (\hat{A} X_t^{(0)} W_f) + H_{t-1}U_f + b_f), \\
	& I_t = {\rm sigmoid}(\sigma (\hat{A} X_t^{(0)} W_i) + H_{t-1} U_i + b_i), \\
	& O_t = {\rm sigmoid}(\sigma (\hat{A} X_t^{(0)} W_o) + H_{t-1} U_o + b_o), \\
	& G_t = {\rm tanh}(\sigma (\hat{A} X_t^{(0)} W_g) + H_{t-1} U_g + b_g), \\
	& C_t = F_t \odot C_{t-1} + I_t \odot G_t, \\
	& H_t = O_t \odot {\rm tanh}(C_t),
	\end{split}
	\label{eq_LGCN}
	\end{equation}
	where ${\rm sigmoid}$ and ${\rm tanh}$ are activation functions, and $\odot$ denotes the Hadamard product.
	We dub the model of Eq.~(\ref{eq_LGCN}) as LGCN.

	\subsubsection{Dual-Attention Method}
	We take the hidden state $H_t$ as the embedding of $G_t$, then the node embeddings of the continuous pose graph $G$ can be denoted as $H=\{H_t | t=1,2,\cdots,T\}$.
	We aim to obtain a feature vector $f^p$ from $H$ to represent the pedestrian pose.
	In this paper, we proposed a dual-attention method, i.e., node-attention and time-attention, to convert the node embeddings $H$ to the pose feature $f^p$.
	
	The node-attention method first performs the average pooling on the time-dimension of $H$:
	\begin{equation}
	\bar{H} = \frac{1}{T} \sum_{t=1}^T H_t ,
	\label{eq_na1}
	\end{equation}
	where 
	$\bar{H} = \{\bar{h}_j | j=1,2,\cdots,14\}$, and $\bar{h}_j \in R^{1\times n}$ denotes the embedding of the $j$-th node.
	Then it learns the weights $a_j^{(n)}$ of each node with the self-attention mechanism: 
	\begin{equation}
	a_j^{(n)} = {\rm softmax}_j \{ \bar{h}_j w^{(n)} \} ,
	\label{eq_na2}
	\end{equation}
	last aggregates node features with $a_j^{(n)}$:
	\begin{equation}
	h^{(n)} = \sum_{j=1}^{14} a_j^{(n)} \bar{h}_j,
	\label{eq_na3}
	\end{equation}
	where $w^{(n)} \in R^{n \times 1}$ is a learnable parameter.
	
	Similar with the node-attention method, the time-attention method can be divided three steps:
	it first performs the average pooling on $H_t=\{h_{tj} | j=1,2,\cdots,14 \}$:
	\begin{equation}
	\bar{h}_t = \frac{1}{14} \sum_{j=1}^{14} h_{tj} ,
	\label{eq_nt1}
	\end{equation}
	then learns the weights of each frame via the self-attention mechanism:
	\begin{equation}
	a_t^{(t)} = {\rm softmax}_t \{ \bar{h}_t w^{(t)} \} ,
	\label{eq_nt2}
	\end{equation}
	finally aggregates the temporal features with the learned weights:
	\begin{equation}
	h^{(t)} = \sum_{t=1}^{T} a_t^{(t)} \bar{h}_t,
	\label{eq_nt3}
	\end{equation}
	where $w^{(t)} \in R^{n \times 1}$ is a learnable parameter.

	The rationale behind the node-attention is that different keypoints in each video clips correspond to different importances; for instance, occluded keypoints may be less important than visible ones.
	While the rationale behind the time-attention is that the occluded frames may contribute less to the retrieval results.
	In conclusion, we learn the feature $h^{(n)}$ and $h^{(t)}$ by the node-attention and time-attention, respectively.
	Finally, we concatenate $h^{(n)}$ and $h^{(t)}$ as the pose feature $f^p$ of the input video as shown in Fig.~\ref{fig_framework}.

	\subsection{Model Training}
	\label{MT}
	Given a video clip $V_i$ corresponding to a label $y_i$, we can learn an appearance feature $f^a_i$ and a pose feature $f^p_i$ by the appearance model and RGCN or LGCN model, respectively.
	Following the existing ReID methods~\cite{BoT,MGH}, we train our model with the triplet loss and identity loss.
	For the triplet loss, we train $f^a_i$ and $f^p_i$ separately:
	\begin{equation}
	\begin{split}
	L_{tri}^a = \sum_{i=1}^{N} \max(0, \delta & + \max \limits_{\substack{y_i=y_j}} \{ \| f_i^a - f_j^a \|_2 \} \\
	& -\min \limits_{\substack{y_i\neq y_k}} \{ \| f_i^a - f_k^a \|_2 \}),
	\end{split}
	\label{eq_triplet_a}
	\end{equation}
	\begin{equation}
	\begin{split}
	L_{tri}^p = \sum_{i=1}^{N} \max(0, \delta & + \max \limits_{\substack{y_i=y_j}} \{ \| f_i^p - f_j^p \|_2 \} \\
	& -\min \limits_{\substack{y_i\neq y_k}} \{ \| f_i^p - f_k^p \|_2 \}),
	\end{split}
	\label{eq_triplet_p}
	\end{equation}
	where $\delta$ is a pre-defined margin and $N$ is the training batch size.
	We then concatenate $f^a_i$ and $f^p_i$ as the final feature $f_i$, and define the identity loss as follows:
	\begin{equation}
	L_{id} = -\frac{1}{N} \sum_{i=1}^{N} \log(\frac{\exp(z_i)}{\sum_{i=1}^{N} \exp(z_i)}) .
	\label{eq_identity_loss}
	\end{equation}
	In Eq.~(\ref{eq_identity_loss}), $z_i=FC(f_i)$ is the classification logits, where $FC$ refers to a fully-connected layer.
	
	In this section, we propose an adaptive training strategy to optimize $L_{tri}^a$ and $L_{tri}^p$, whose purpose is to impose a larger punishment to the larger loss.
	To this end, we define an adaptive parameter $\lambda$:
	\begin{equation}
	\lambda = L_{tri}^a / (L_{tri}^a + L_{tri}^p).
	\label{eq_lamda}
	\end{equation}
	Then the final loss can be denoted as the linear combination of above-mentioned losses:
	\begin{equation}
	L = L_{id} + \lambda L_{tri}^a + (1-\lambda) L_{tri}^p.
	\label{eq_loss}
	\end{equation}
	By minimizing $L$, we could jointly learn the discriminative appearance feature $f^a_i$ and pose feature $f^p_i$.

	\begin{table}[h]
	\small
	\centering
	\caption{Statistics of three datasets in our experiment.}
	\begin{tabular}{c|cccc}
		\hline
		Dataset   & Identities & Videos & Camera  & Splits    \\ \hline
		Mars      & 1,261       & 20,751  & 6     & 625/636   \\
		DukeMTMC  & 1,404       & 4,832   & 8     & 702/702   \\
		iLIDS-VID & 300        & 600    & 2       & 702/702   \\ \hline
	\end{tabular}
	\label{table_dataset}
\end{table}
	
	\section{Experiments}
	\label{experiment}
	
	\subsection{Datasets and Experimental Implementations}
	\textbf{Datasets.}
	To validate the effectiveness of the learned pose feature in video-based ReID, we conduct our experiments on three widely-used datasets: Mars~\cite{Mars}, DukeMTMC~\cite{duke1,duke2} and iLIDS-VID~\cite{ilids}.
	Table~\ref{table_dataset} summarizes the statistics of three datasets.
	For the evaluation metrics, we report the mean Average Precision (mAP) and Cumulative Match Characteristic (CMC).

	\begin{table}[t]
		\centering
		\caption{Comparison with the appearance models.}
		\begin{tabular}{c|cc|cc}
			\hline
			\multirow{2}{*}{Methods} & \multicolumn{2}{c|}{Mars} & \multicolumn{2}{c}{iLIDS-VID} \\ \cline{2-5} 
			& mAP         & Rank1       & mAP          & Rank1          \\ \hline
			AP                      & 81.1\%        & 85.2\%        & 87.6\%             & 82.0\%               \\
			AP+RGCN                 & 82.6\% $\uparrow$            & 87.9\% $\uparrow$           & 89.4\% $\uparrow$             & 84.7\% $\uparrow$               \\ 
			AP+LGCN                 & 82.3\% $\uparrow$            & 88.0\% $\uparrow$           & 90.4\% $\uparrow$             & 86.0\% $\uparrow$               \\ \hline
			AA                      & 73.0\%             & 80.8\%            & 85.2\%              & 78.7\%                \\
			AA+RGCN                 & 76.4\% $\uparrow$             & 83.6\% $\uparrow$             & 87.3\% $\uparrow$            & 81.3\% $\uparrow$              \\ 
			AA+LGCN                 & 76.4\% $\uparrow$            & 83.5\% $\uparrow$             & 87.8\% $\uparrow$            & 80.7\% $\uparrow$              \\ \hline
			RA                      & 80.7\%            & 85.5\%            & 87.0\%              & 80.7\%                \\
			RA+RGCN                 & 81.8\% $\uparrow$            & 86.8\% $\uparrow$            &  88.7\% $\uparrow$             &  83.3\% $\uparrow$               \\ 
			RA+LGCN                 & 81.9\% $\uparrow$            & 87.5\% $\uparrow$            &  89.3\% $\uparrow$             &  84.7\% $\uparrow$               \\ \hline
			BiC                       & 86.0\%            & 90.2\%            & 91.1\%             & 88.3\%               \\
			BiC+RGCN                  & 86.0\%            & 90.7\% $\uparrow$            & 91.5\% $\uparrow$             & 89.6\% $\uparrow$               \\ 
			BiC+LGCN                  & 86.5\% $\uparrow$            & 91.1\% $\uparrow$            & 91.6\% $\uparrow$             & 90.2\% $\uparrow$               \\ \hline
		\end{tabular}
		\label{table_appearence}
	\end{table}
	
	\begin{table}[t]
		\centering
		\caption{Comparison with the other temporal GCN models on Mars~\cite{Mars}.}
		\begin{tabular}{c|cc|c}
			\hline
			\multirow{2}{*}{Methods} & \multicolumn{2}{c|}{Mars} & \multirow{2}{*}{Training Time} \\ \cline{2-3}
			& mAP         & Rank1       &                                \\ \hline
			AP                      & 81.1\%      & 85.2\%      & 28.30 hours                               \\
			AP+RGCN                 & 82.6\%            & 87.9\%            & 32.14 hours                               \\
			AP+GCN\&RNN             & 82.3\%            & 87.7\%            & 41.55 hours                               \\
			AP+LGCN                 & 82.3\%            & 88.0\%            & 38.82 hours                               \\
			AP+GCN\&LSTM            & 81.8\%            & 86.1\%            & 47.60 hours                               \\ 
			AP+ST-GCN~\cite{ST-GCN}            & 82.3\%            & 86.6\%            & 32.25 hours                               \\
			AP+EvolveGCN~\cite{Evolvegcn}            & 81.7\%            & 86.8\%            & 37.36 hours                               \\ \hline
		\end{tabular}
		\label{table_dygcn_mars}
	\end{table}
	
	\begin{table}[t]
	\centering
	\caption{Comparison with the other temporal GCN models on iLIDS-VID~\cite{ilids}.}
	\begin{tabular}{c|cc|c}
		\hline
		\multirow{2}{*}{Methods} & \multicolumn{2}{c|}{iLIDS-VID} & \multirow{2}{*}{Training Time} \\ \cline{2-3}
		& mAP         & Rank1       &                                \\ \hline
		AP                      & 87.6\%      & 82.0\%      & 3.00 hours                               \\
		AP+RGCN                 & 89.4\%            & 84.7\%            & 3.25 hours                               \\
		AP+GCN\&RNN             & 88.5\%            & 84.0\%            & 4.30 hours                               \\
		AP+LGCN                 & 90.4\%            & 86.0\%            & 3.33 hours                               \\
		AP+GCN\&LSTM            & 90.3\%            & 85.8\%            & 4.23 hours                               \\ 
		AP+ST-GCN~\cite{ST-GCN}            & 89.3\%            & 84.0\%            & 3.22 hours                               \\
		AP+EvolveGCN~\cite{Evolvegcn}            & 86.0\%            & 81.3\%            & 3.30 hours                                \\ \hline
	\end{tabular}
	\label{table_dygcn_ilids}
\end{table}

	\begin{table}[t]
	\centering
	\caption{Validation of the dual-attention mechanism.}
	\begin{tabular}{c|cc|cc}
		\hline
		\multirow{2}{*}{Methods} & \multicolumn{2}{c|}{Mars} & \multicolumn{2}{c}{iLIDS-VID} \\ \cline{2-5} 
		& mAP         & Rank1       & mAP          & Rank1          \\ \hline
		RGCN+mean                 & 82.1\%             & 86.3\%            & 88.6\%              & 83.3\%              \\
		RGCN+TAM                 & 82.4\%             & 87.1\%            & 88.3\%              & 82.3\%              \\
		RGCN+NAM                 & 82.2\%             & 86.9\%            & 88.6\%              & 84.0\%              \\
		RGCN+DAM                 & 82.6\%             & 87.9\%            & 89.4\%              & 84.7\%              \\ \hline
		LGCN+mean                 & 81.5\%             & 86.4\%            & 87.7\%              & 82.3\%              \\
		LGCN+TAM                 & 81.7\%             & 86.8\%            & 90.2\%              & 84.5\%              \\
		LGCN+NAM                 & 81.6\%             & 85.9\%            & 89.0\%              & 84.7\%              \\
		LGCN+DAM                 & 82.3\%             & 88.0\%            & 90.4\%              & 86.0\%              \\ \hline
	\end{tabular}
	\label{table_dam}
\end{table}

\textbf{Implementations.}
We adopt the ResNet50~\cite{ResNet} as the feature extractor.
We train our model 800 times with the Adam optimizer~\cite{Adam}.
The learning rate is initialized to $3 \times 10^{-4}$, and decays by 0.1 every 200 epochs.
In each training epoch, we select 32 video sequences from 8 identities, while the length of each video sequence is set to 10.
The dimension of the appearance feature $f^a$ is set to 2048, while that of the pose feature $f^p$ is set to 512.

\subsection{Ablation Studies}
In this section, we conduct our ablation studies on Mars~\cite{Mars} and iLIDS-VID~\cite{ilids}; we report mAP and Rank1 of CMC for comparison.

\subsubsection{Comparison with the Appearance Models}
In this section, we compare our method with multiple appearance models, including the average pooling (AP), attention aggregation (AA), RNN aggregation (RA) and BiCnet-TKS~\cite{BiCnet}.
The equations of the first three aggregators, i.e., AP, AA and RA, are presented in Section~\ref{AFL}, while BiCnet-TKS (BiC for short) is the most recent state-of-the-arts appearance model for video-based ReID.
We implement the RGCN and LGCN with one graph convolutional layer.

We report the experimental results in Table~\ref{table_appearence}, from which we could draw the following conclusions:
1) both the RGCN and LGCN outperform the baseline appearance model, which demonstrates the effectiveness of the pose feature in video-based ReID;
2) compared with RGCN model, LGCN could achieve higher accuracy in most cases.

\subsubsection{Comparison with Other Temporal GCN Models}
In this section, we compare our RGCN and LGCN model with other temporal GCN models, including GCN\&RNN pipeline, GCN\&LSTM pipeline, ST-GCN~\cite{ST-GCN} and EvolveGCN~\cite{Evolvegcn}.
The GCN\&RNN pipeline is illustrated in Eq.~(\ref{eq_GCN}) and Eq.~(\ref{eq_RNN}), while the GCN\&LSTM pipeline could be implemented by combining Eq.~(\ref{eq_GCN}) and LSTM model.
ST-GCN~\cite{ST-GCN} is a popular framework for temporal skeleton feature learning.
EvolveGCN~\cite{Evolvegcn} is recently proposed for dynamic graph representation learning.

We use two Tesla V100 GPU to train Mars and only use one to train iLIDS-VID.
The experimental results and training time on Mars and iLIDS-VID are reported in Table~\ref{table_dygcn_mars} and Table~\ref{table_dygcn_ilids}, respectively.
As can be seen, even though the GCN\&RNN (GCN\&LSTM) could achieve similar performance than RGCN (LGCN), it costs more time for model training.
Otherwise, our model outperforms both ST-GCN and EvolveGCN on Mars and iLIDS-VID.

\begin{figure}[t]
	\centering
	\subfigure[RGCN]{\includegraphics[width=0.232\textwidth]{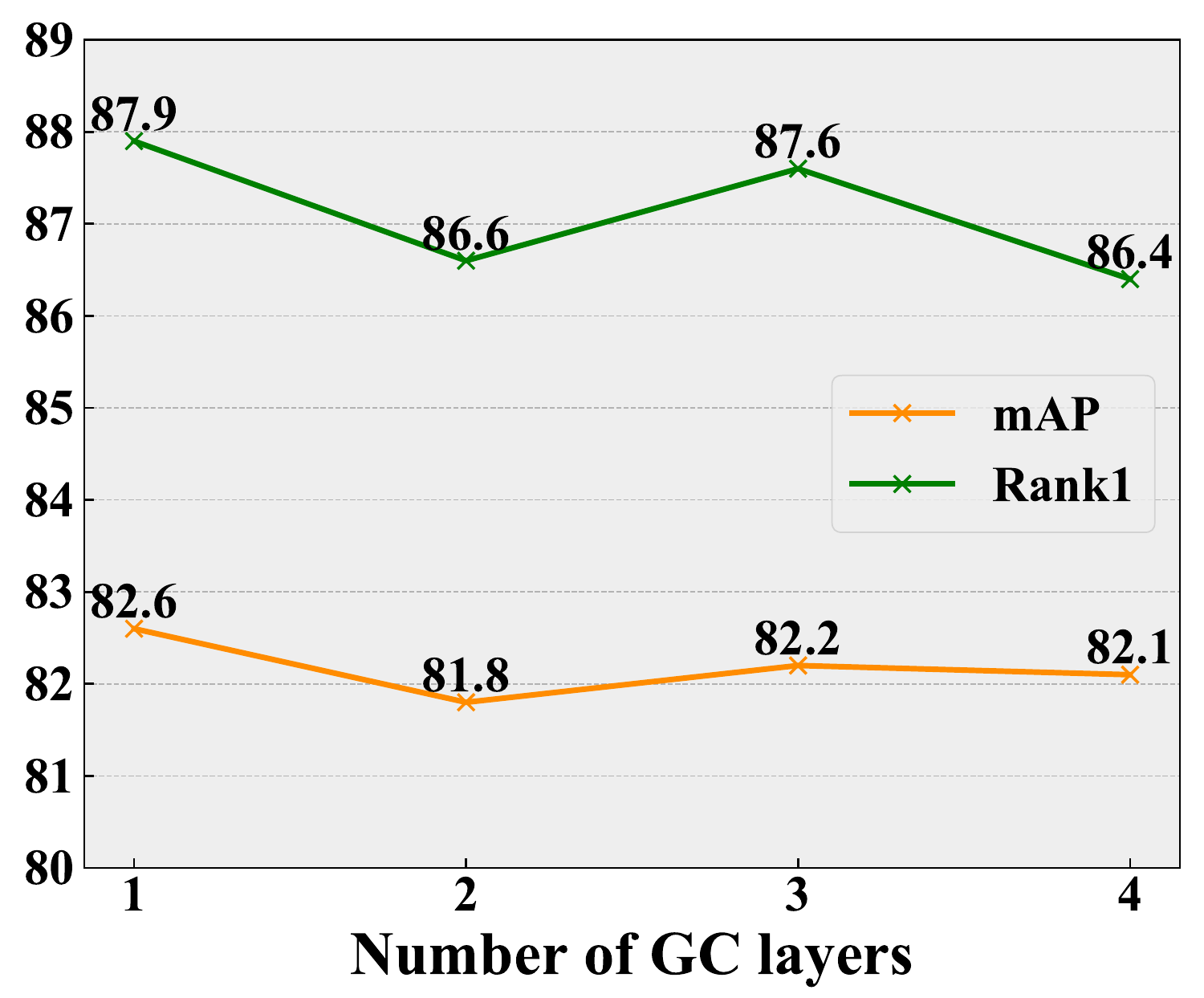}}
	\subfigure[LGCN]{\includegraphics[width=0.232\textwidth]{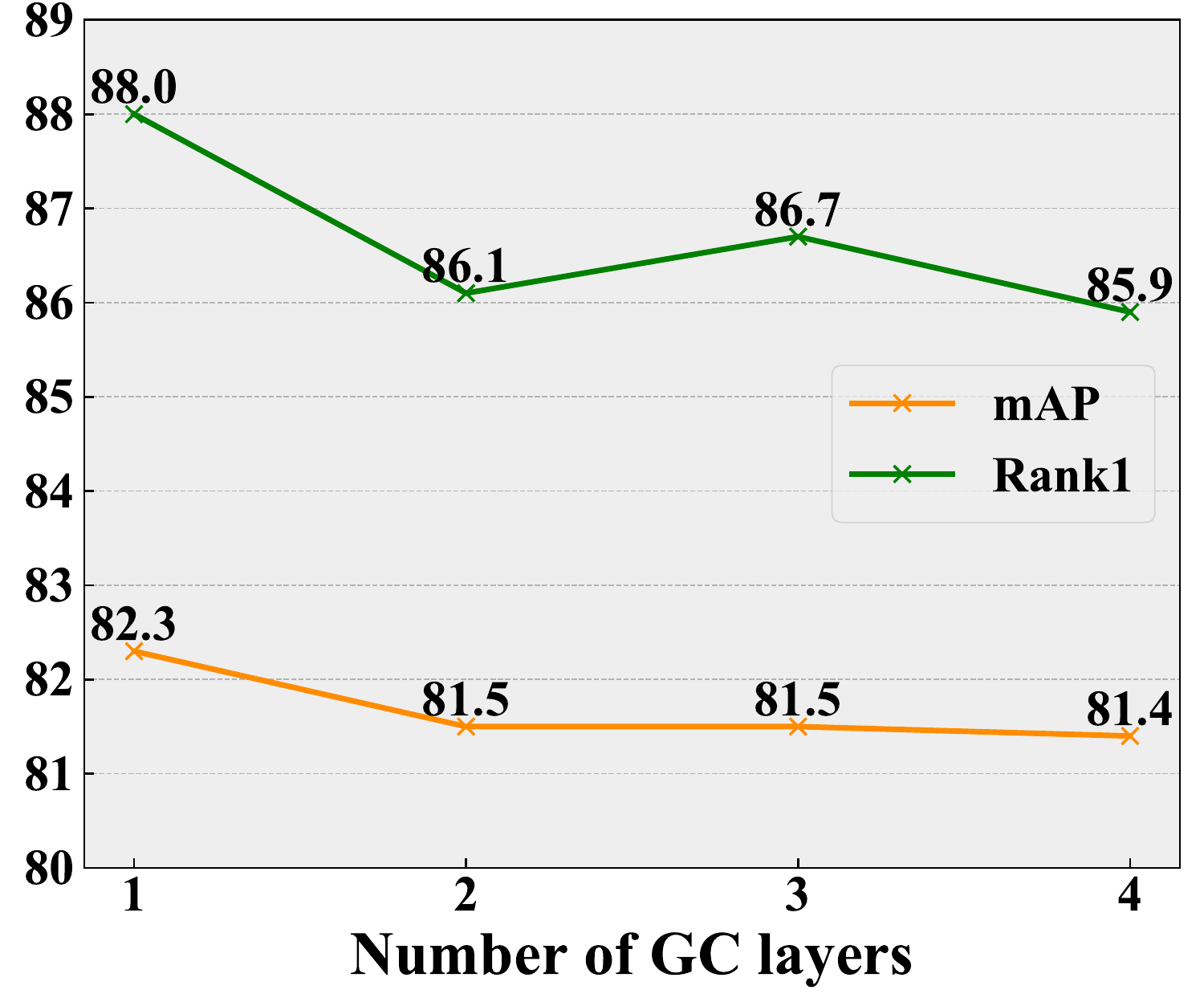}}
	\caption{Performance of RGCN and LGCN with different GC layers on Mars~\cite{Mars}.
		Numbers shown in the figure are percentages.}
	\label{fig_mars_gc_layer}
\end{figure}

\begin{figure}[h]
	\centering
	\subfigure[RGCN]{\includegraphics[width=0.232\textwidth]{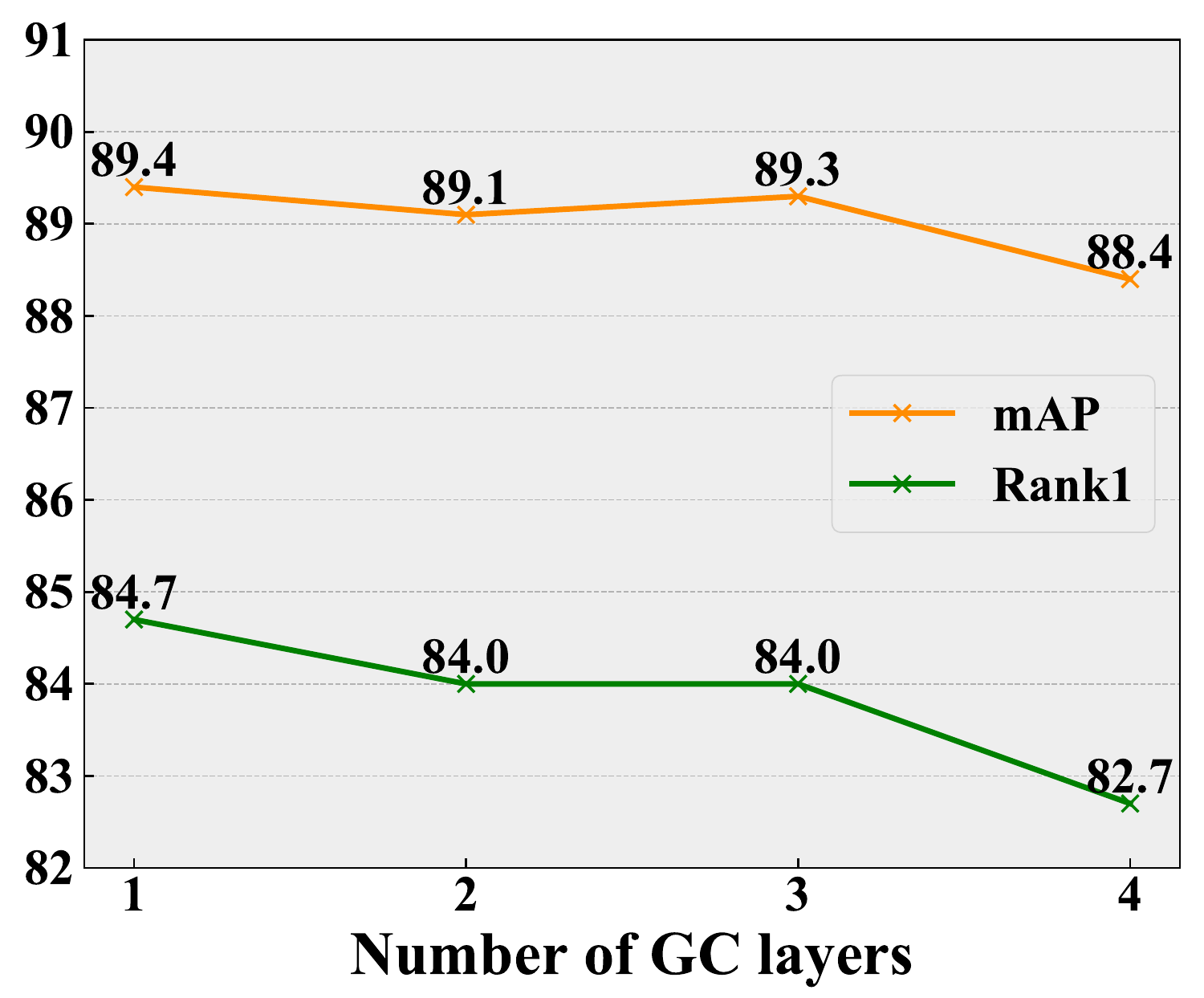}}
	\subfigure[LGCN]{\includegraphics[width=0.232\textwidth]{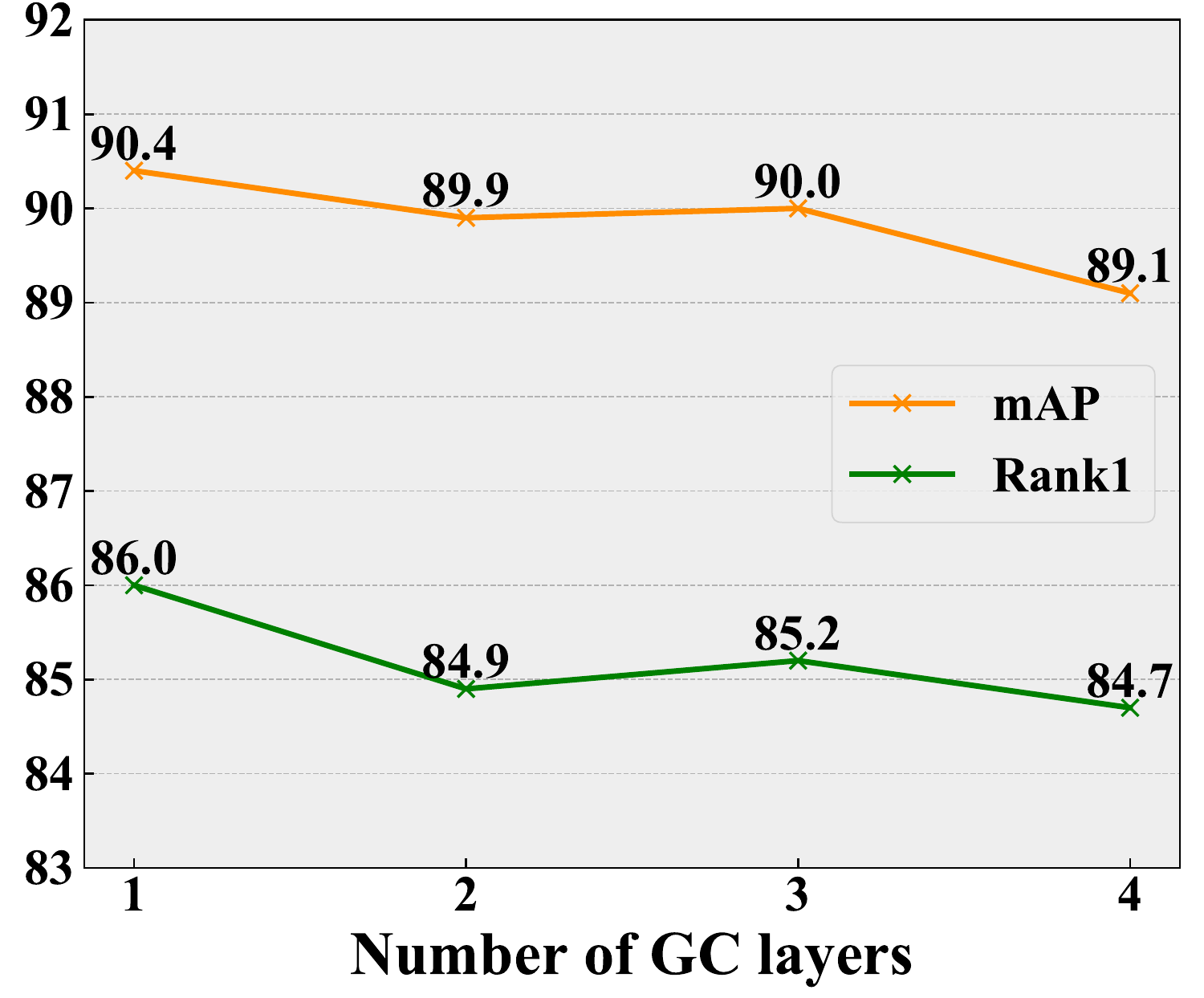}}
	\caption{Performance of RGCN and LGCN with different GC layers on iLIDS-VID~\cite{ilids}.
		Numbers shown in the figure are percentages.}
	\label{fig_ilids_gc_layer}
\end{figure}

\begin{figure}[t]
	\centering
	\subfigure[RGCN]{\includegraphics[width=0.232\textwidth]{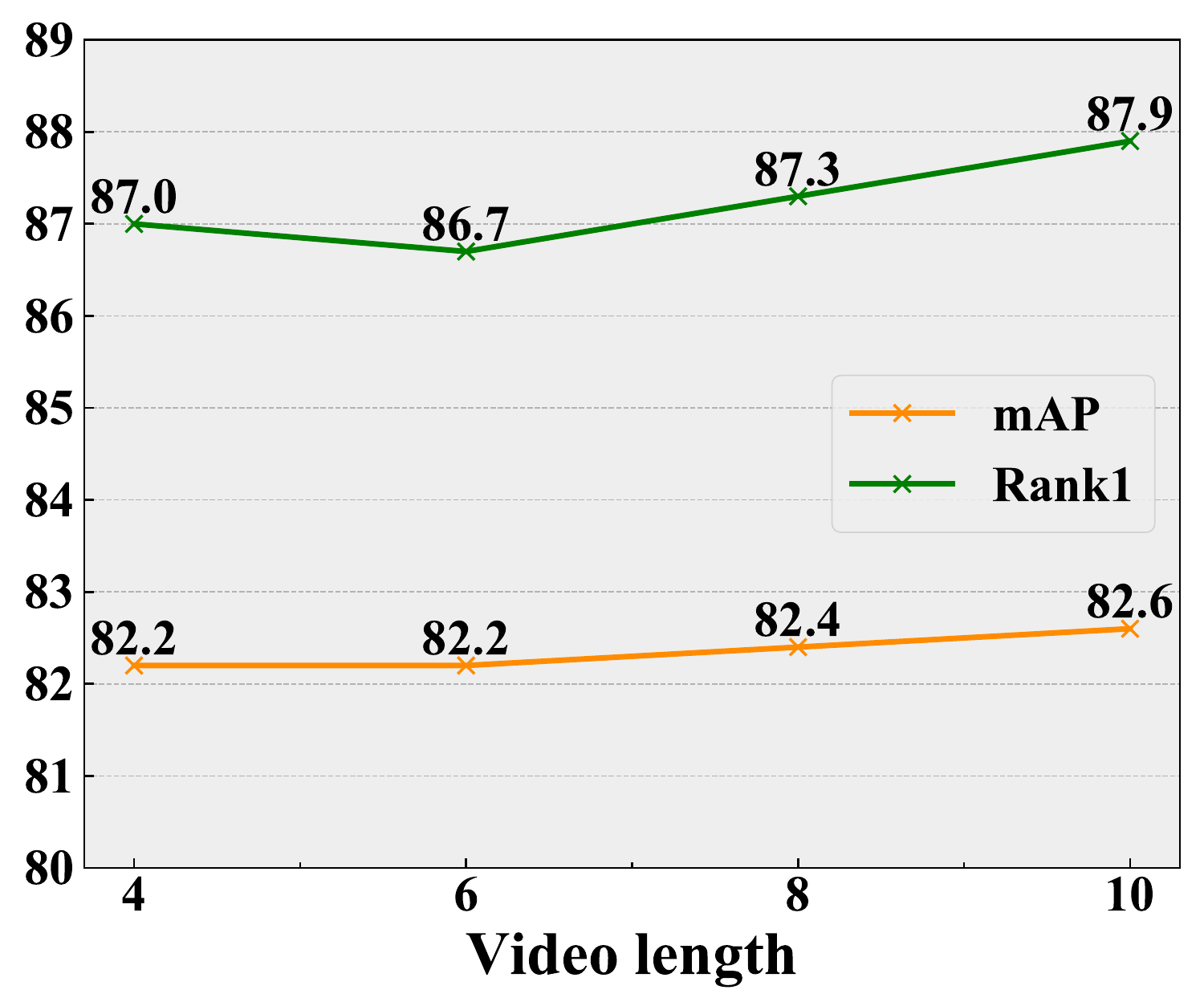}}
	\subfigure[LGCN]{\includegraphics[width=0.232\textwidth]{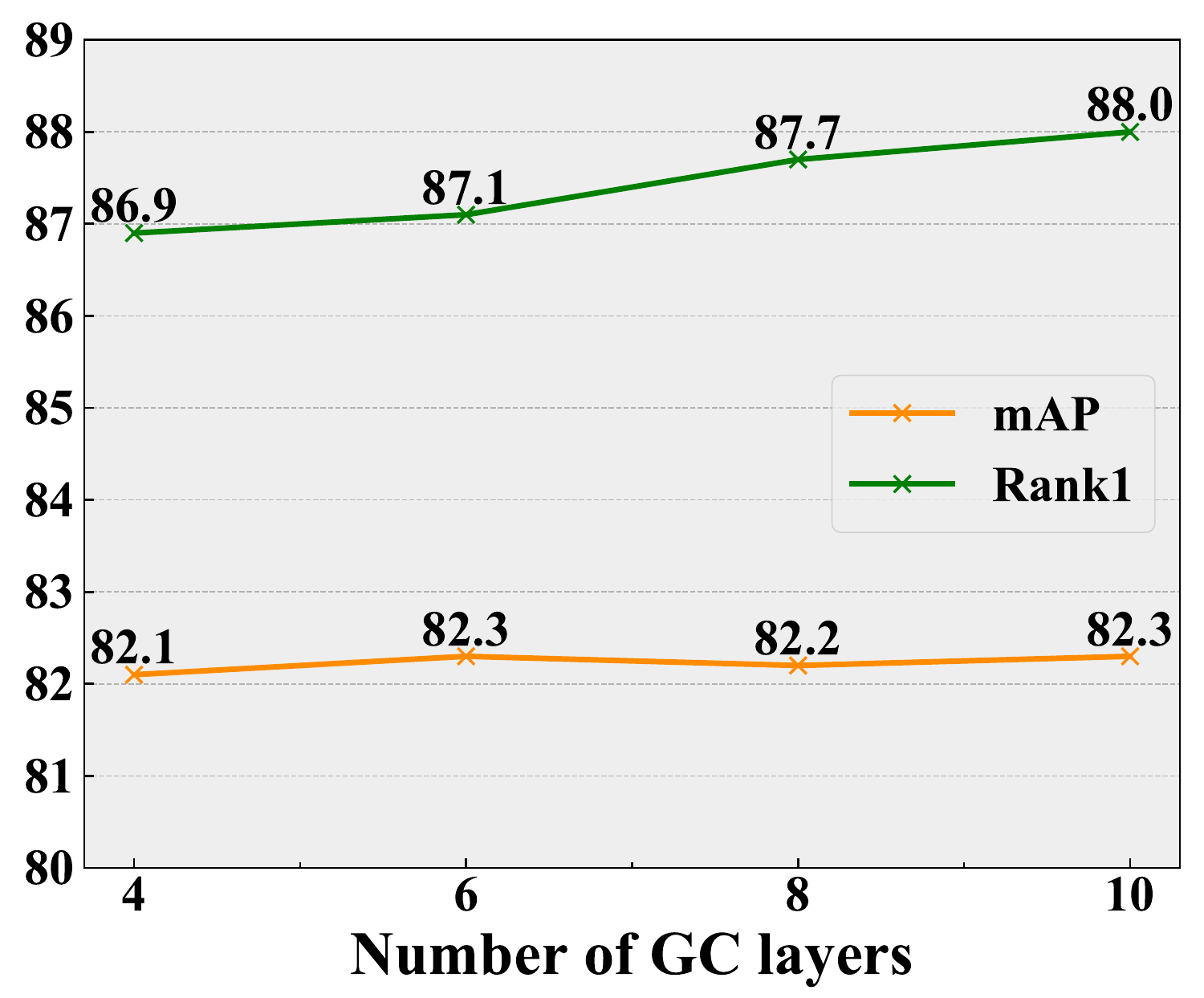}}
	\caption{Performance of RGCN and LGCN with different video lengths $T$ on Mars~\cite{Mars}.
		Numbers shown in the figure are percentages.}
	\label{fig_mars_T}
\end{figure}

\begin{figure}[h]
	\centering
	\subfigure[RGCN]{\includegraphics[width=0.232\textwidth]{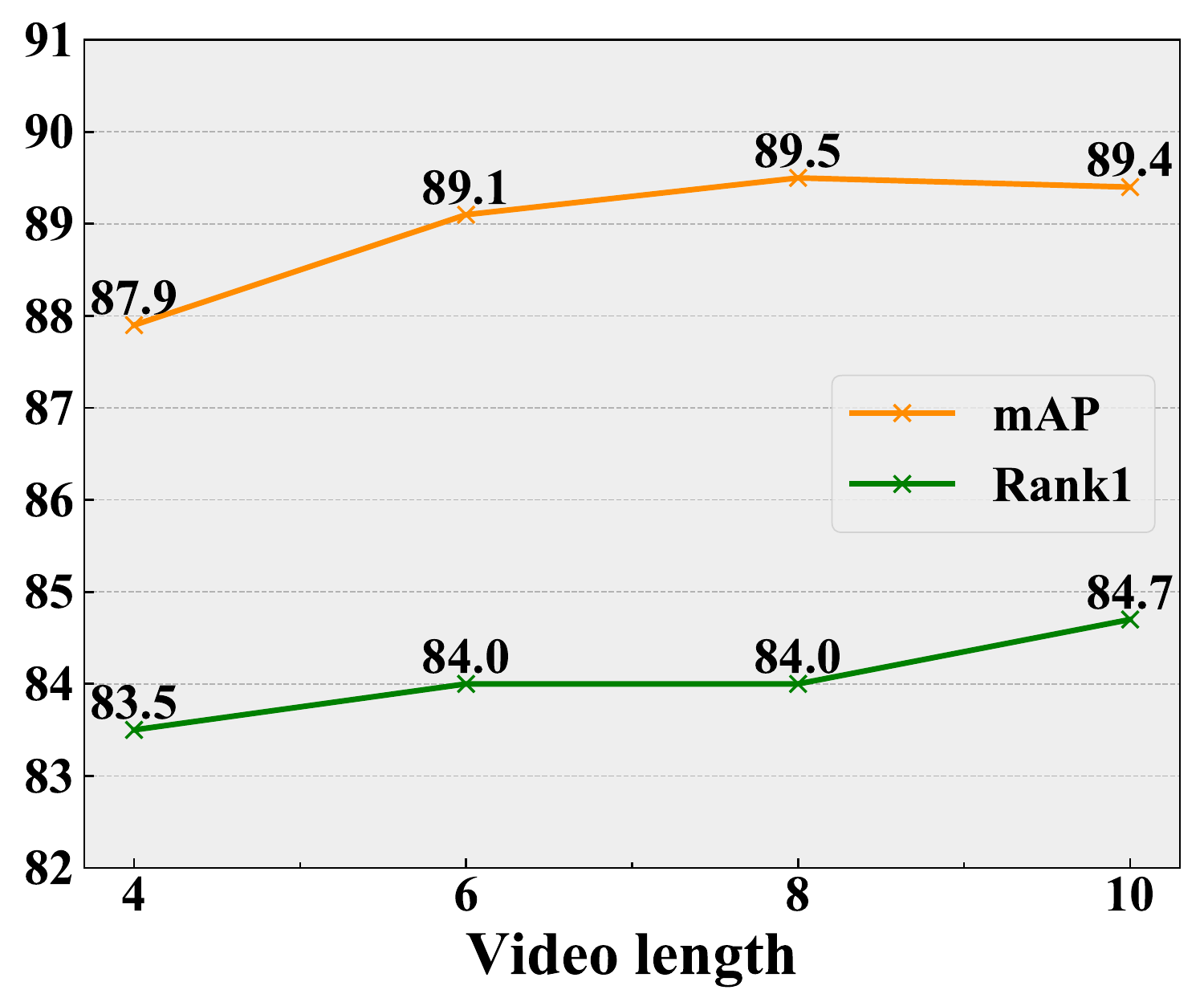}}
	\subfigure[LGCN]{\includegraphics[width=0.232\textwidth]{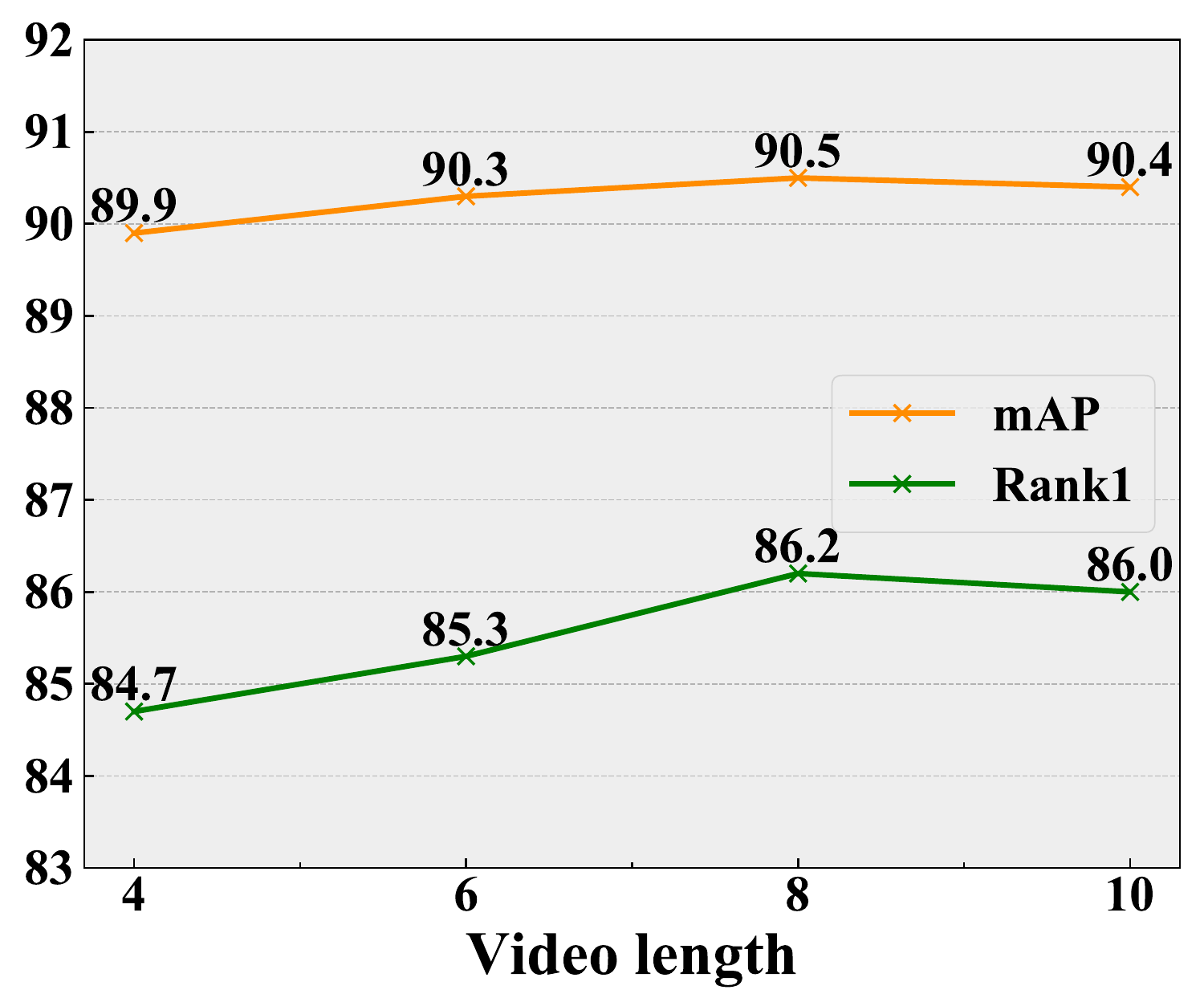}}
	\caption{Performance of RGCN and LGCN with different video lengths $T$ on iLIDS-VID~\cite{ilids}. Numbers shown in the figure are percentages.}
	\label{fig_ilids_T}
\end{figure}

\subsubsection{Validation of the Dual-Attention Method}
In this section, we validate the effectiveness of the dual-attention method (DAM).
Specifically, we compare our DAM with mean pooling, time-attention method (TAM) and node-attention method (NAM): mean pooling takes the mean value of the node embeddings as the pose feature; TAM and NAM obtain graph representation with time-level attention and node-level attention, respectively.
As shown in Table~\ref{table_dam}, the attention method can improve the performance of the mean pooling, while our DAM outperforms TAM and NAM.

\begin{table}[h]
	\small
	\centering
	\caption{Comparison with the different values of $\lambda$.}
	\begin{tabular}{c|c|cc|cc}
		\hline
		\multirow{2}{*}{Method} & \multirow{2}{*}{$\lambda$} & \multicolumn{2}{c|}{Mars} & \multicolumn{2}{c}{iLIDS-VID} \\ \cline{3-6} 
		&                        & mAP        & Rank1        & mAP        & Rank1        \\ \hline
		\multirow{3}{*}{RGCN}   & Adaptive               & 82.6\%           & 87.9\%             & 89.4\%           & 84.7\%             \\
		& 0.5                    & 82.2\%$\downarrow$           & 86.6\%$\downarrow$             & 88.7\%$\downarrow$           & 84.0\%$\downarrow$             \\
		& 1.0                    & 82.1\%$\downarrow$           & 86.8\%$\downarrow$             & 87.5\%$\downarrow$           & 82.0\%$\downarrow$             \\ \hline
		\multirow{3}{*}{LGCN}   & Adaptive               & 82.3\%           & 88.0\%             & 90.4\%           & 86.0\%             \\
		& 0.5                    & 82.0\%$\downarrow$           & 87.1\%$\downarrow$             & 89.5\%$\downarrow$           & 84.7\%$\downarrow$              \\
		& 1.0                    & 81.7\%$\downarrow$           & 87.3\%$\downarrow$             & 89.3\%$\downarrow$           & 84.3\%$\downarrow$             \\ \hline
	\end{tabular}
	\label{table_lamda}
\end{table}

\subsubsection{Impact of Graph Convolutional Layers}
In this section, we test the performance of RGCN and LGCN with different graph convolutional (GC) layers.
Specifically, we choose the average pooling (AP) as the baseline model and implement RGCN and LGCN with one to four GC layers.
The experimental results on Mars and iLIDS-VID are reported in Fig.~\ref{fig_mars_gc_layer} and Fig.~\ref{fig_ilids_gc_layer}, respectively.
As can be seen, the accuracy decreases as the number of GC layers increases.
In our method, each pose graph only contains 14 nodes, which would result the over-smoothing~\cite{low_pass,low_pass2} after multiple GC layers.

\subsubsection{Impact of the Sequence Length}
Limited by the GPU memory, we input a video clip with a specific length $T$ for each iteration.
In this section, we explore the retrieval accuracy under different $T$.
Specifically, we verify four values for $T$: 4, 6, 8 and 10.
We choose the average pooling (AP) as the baseline model and implement RGCN and LGCN with The retrieval results on Mars and iLIDS-VID are presented in Fig.~\ref{fig_mars_T} and Fig.~\ref{fig_ilids_T}, respectively.
As can be seen, the overall tendency is that the larger $T$ contributes to the higher accuracy.

\begin{figure}[h]
	\centering
	\subfigure[RGCN]{\includegraphics[width=0.232\textwidth]{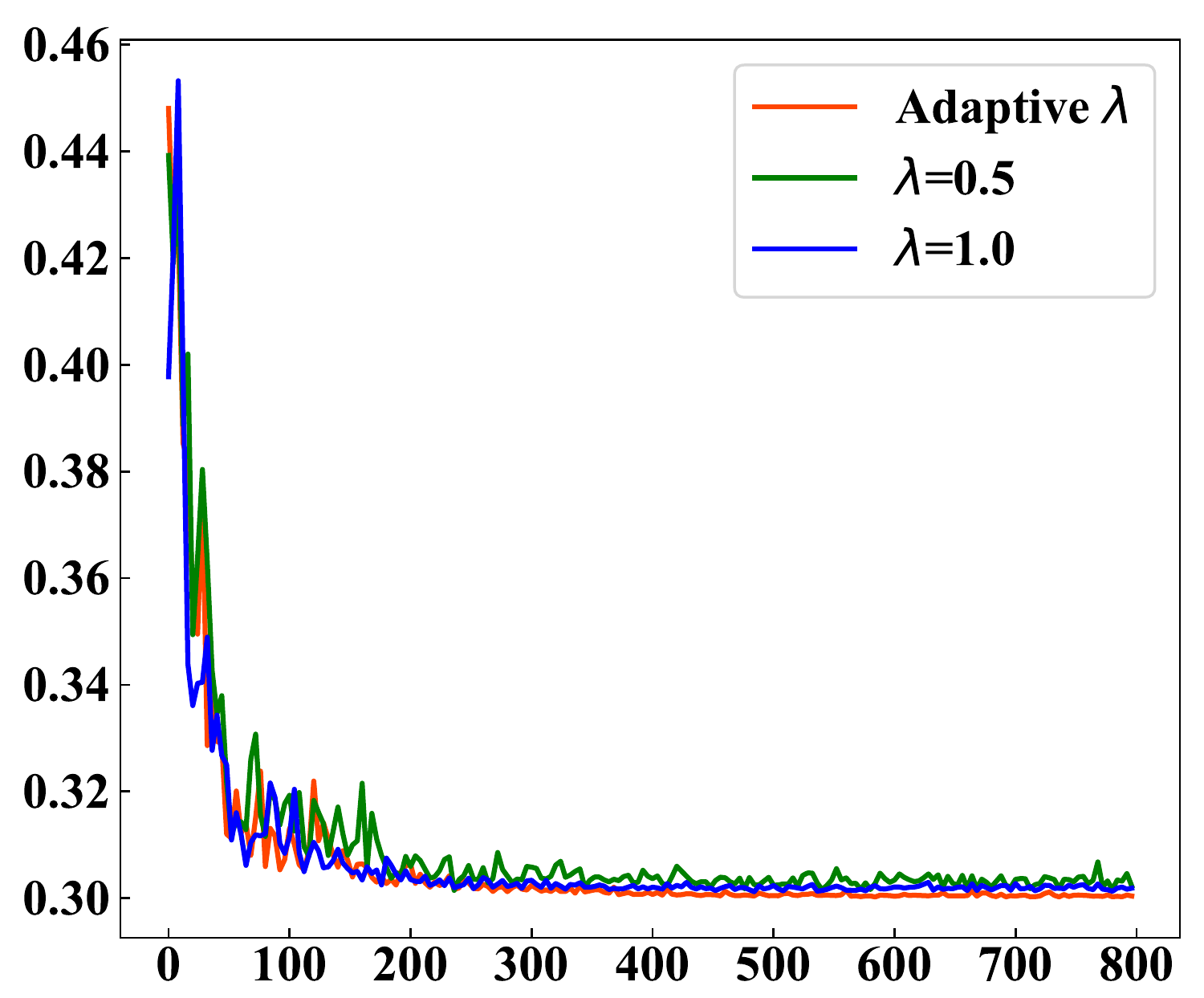}}
	\subfigure[LGCN]{\includegraphics[width=0.232\textwidth]{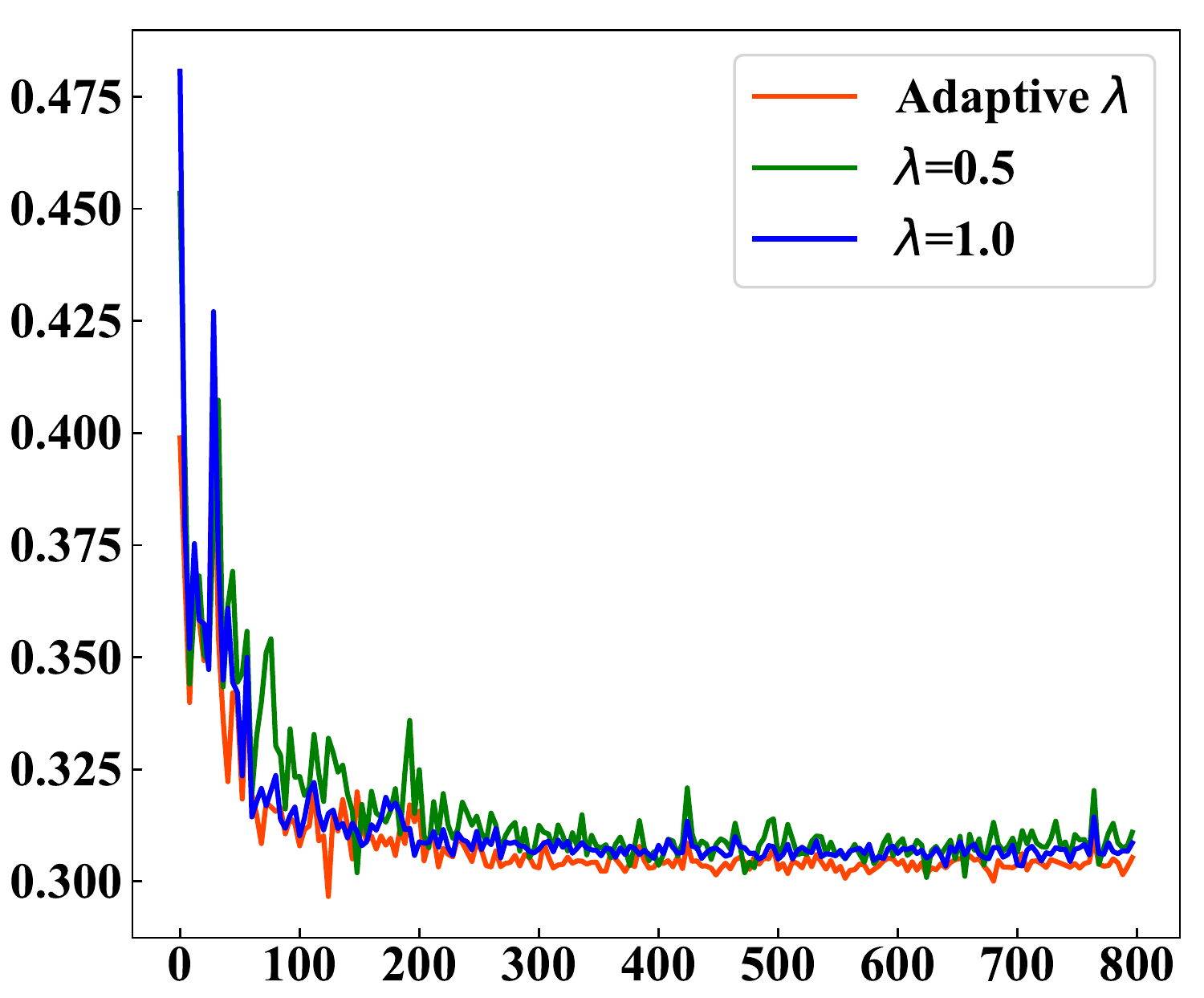}}
	\caption{The pose triplet loss $L_{tri}^p$ under the adaptive $\lambda$ and fixed values (0.5 and 1.0) on iLIDS-VID~\cite{ilids}. The abscissa and ordinate indicate the training epoch and loss value, respectively.}
	\label{fig_pose_loss}
\end{figure}

	\begin{table*}[t]
	\small
	\centering
	\caption{Comparison with the state-of-the-arts. The three best scores are indicated in \textcolor{red}{red}, \textcolor{blue}{blue}, \textcolor{green}{green}, respectively.}
	\begin{tabular}{c|c|cccc|cccc|ccc}
		\hline
		\multirow{2}{*}{Methods} & \multirow{2}{*}{Venue} & \multicolumn{4}{c|}{Mars}     & \multicolumn{4}{c|}{DukeMTMC} & \multicolumn{3}{c}{iLIDS-VID} \\ \cline{3-13} 
		&                        & mAP  & Rank1 & Rank5 & Rank20 & mAP  & Rank1 & Rank5 & Rank20 & Rank1    & Rank5   & Rank20   \\ \hline
		RGSA~\cite{RGSA}	& AAAI2020    & 84.0\%     & 89.4\%      & -      & -       & 95.8\%     & 97.2\%       & -      & -       & 86.0\%         & 98.0\%        & 99.4\%         \\
		FGRA~\cite{FGRA}	& AAAI2020    & 81.2\%     & 87.3\%      & 96.0\%      & 98.0\%       & -     & -       & -      & -       & 88.0\%         & 96.7\%        & 99.3\%         \\
		MGH~\cite{MGH} & CVPR2020   & 85.8\%     & 90.0\%      & 96.7\%      & \textcolor{green}{98.5\%}       & -     & -      & -      & -       & 85.6\%         & 97.1\%        & \textcolor{blue}{99.8\%}         \\
		STGCN~\cite{STGCN} & CVPR2020   & 83.7\%     & 90.0\%      & 96.4\%      & 98.3\%       & 95.7\%     & \textcolor{blue}{97.3\%}      & \textcolor{red}{99.3\%}      & \textcolor{red}{99.7\%}       & -         & -        & -         \\
		TCLNet~\cite{TCLNet}	& ECCV2020    & 85.8\%     & 89.8\%      & -      & -       & 96.2\%     & 96.9\%       & -      & -       & 86.6\%         & -        & -         \\
		AP3D~\cite{AP3D}	& ECCV2020    & 85.1\%     & 90.1\%      & -      & -       & 95.6\%     & 96.3\%       & -      & -       & 86.7\%         & -        & -         \\
		ASTA~\cite{ASTA}	& MM2020    & 84.1\%     & 90.4\%      & \textcolor{blue}{97.0\%}      & \textcolor{red}{98.8\%}       & -     & -       & -      & -       & 88.1\%         & \textcolor{red}{98.6\%}        & -         \\
		SSN3D~\cite{SSN3D}	& AAAI2021    & \textcolor{green}{86.2\%}     & 90.1\%      & 96.6\%      & 98.0\%       & \textcolor{green}{96.3\%}     & 96.8\%       & 98.6\%      & \textcolor{blue}{99.4\%}       & 88.9\%         & 97.3\%        & -         \\
		BiCnet~\cite{BiCnet}	& CVPR2021    & 86.0\%     & 90.2\%      & -      & -       & 96.1\%     & 96.3\%       & -      & -       & -         & -        & -         \\
		GRL~\cite{GRL}	& CVPR2021    & 84.8\%     & \textcolor{green}{91.0\%}      & 96.7\%      & 98.4\%       & -     & -       & -      & -       & \textcolor{red}{90.4\%}         & 98.3\%        & \textcolor{blue}{99.8\%}         \\
		CTL~\cite{CTL} & CVPR2021   & \textcolor{red}{86.7\%}     & \textcolor{red}{91.4\%}      & \textcolor{green}{96.8\%}      & \textcolor{green}{98.5\%}       & -     & -      & -      & -       & \textcolor{green}{89.7\%}         & 97.0\%        & \textcolor{red}{100\%}         \\
		STRF\cite{STRF} & ICCV2021                        & 86.1\%     & 90.3\%      & -      & -       & \textcolor{blue}{96.4\%}     & \textcolor{red}{97.4\%}       & -       & -       & 89.3\%         & -        & -         \\
		HMN~\cite{HMN} & TCSVT                        & 82.6\%    & 88.5\%      & 96.2\%      & 98.1\%       & 95.1\%     & 96.3\%       & -       & -       & 83.3\%         & 97.1\%        & 99.5\%         \\
		SGMN~\cite{SGMN} & TCSVT                        & 85.4\%    & 90.8\%      & -      & -       & \textcolor{green}{96.3\%}     & 96.9\%       & -       & -       & 88.7\%         & 96.7\%        & 99.3         \\
		\hline
		ours                     & -                      & \textcolor{blue}{86.5\%} & \textcolor{blue}{91.1\%}  & \textcolor{red}{97.2\%}  & \textcolor{blue}{98.6\%}   & \textcolor{red}{96.5\%} & \textcolor{green}{97.1\%}  & \textcolor{green}{98.8\%}  & \textcolor{red}{99.7\%}   & \textcolor{blue}{90.2\%}     & \textcolor{blue}{98.5\%}    & \textcolor{green}{99.6\%}     \\ \hline
	\end{tabular}
	\label{table_comparison}
\end{table*}

\begin{figure*}[h]
	\centering
	\includegraphics[width=0.9\textwidth]{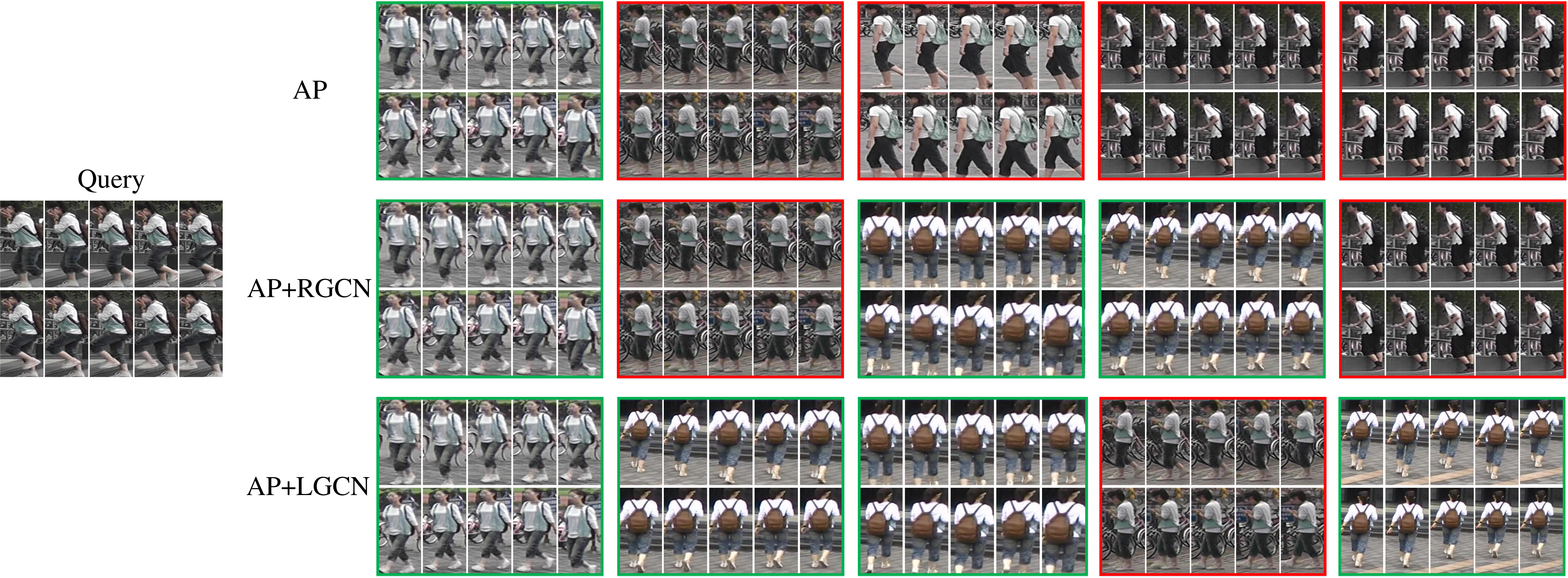}
	\caption{Visualization of the rank-5 retrieval results, where the green box and red box indicate the true positive result and false positive result, respectively.}
	\label{fig_vis}
\end{figure*}

\begin{figure}[]
	\centering
	\subfigure[Time attention scores]{\includegraphics[width=0.48\textwidth]{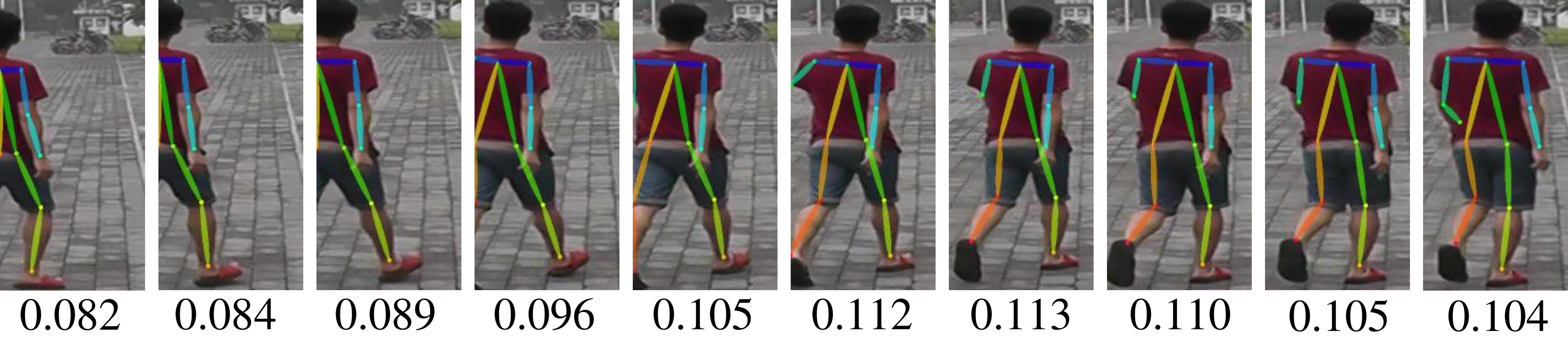}}
	\subfigure[Node attention scores]{\includegraphics[width=0.22\textwidth]{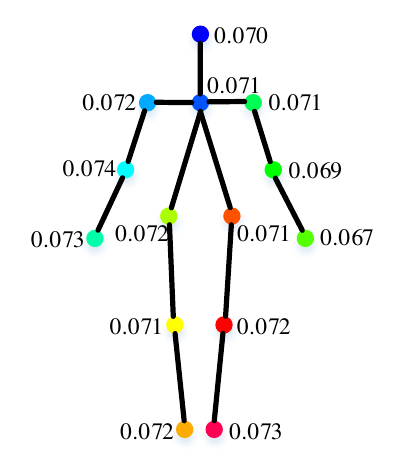}}
	\caption{Visualization of the attention scores of each frame and each node learned by DAM.
	}
	\label{fig_dam}
\end{figure}

\subsubsection{Validation of the Adaptive Loss}
Eq.(~\ref{eq_loss}) defines an adaptive loss by a changeable $\lambda$.
In this section, we aim to compare the performance of the adaptive $\lambda$ and fixed $\lambda$.
We test two fixed values for $\lambda$, i.e., 0.5 and 1.0.
Their respective performances are reported in Table~\ref{table_lamda}.
As can be seen, the adaptive $\lambda$ outperforms the fixed $\lambda$ on both Mars and iLIDS-VID.
Moreover, we visualize the pose triplet loss $L_{tri}^p$ during training iLIDS-VID~\cite{ilids} in Fig.~\ref{fig_pose_loss}.
The conclusion is that the loss defined by adaptive $\lambda$ could converge to a smaller value compared with fixed $\lambda$.

	\subsection{Comparison and Visualization}
	\subsubsection{Comparison with the State-of-the-arts}
	In this section, we compare our method with the state-of-the-art (SOTA) methods, including STRF~\cite{STRF}, GRL~\cite{GRL}, CTL\cite{CTL}, SSN3D~\cite{SSN3D}, TCLNet~\cite{TCLNet}, CPA~\cite{CPA}, and so on.
	We choose BiCnet-TKS~\cite{BiCnet} as our baseline appearance model.
	As shown in Table~\ref{table_comparison}, our method achieves the competitive results on these three datasets.

	\subsubsection{Visualization}
	In this section, we visualize the retrieval results from Mars~\cite{Mars} in  Fig.~\ref{fig_vis}.
	As can be seen, for the AP, the video clips of the second and third columns have the similar appearances with the query video, which results the false positive results.
	While for RGCN, even though the video clips of the third and fourth columns have the different appearances and camera views with the query video, it still achieves the true positive results.
	The similar phenomenon is shown in LGCN.
	Thereby, we could draw the conclusion that the learned pose feature can effectively enlarge the inter-class variance for pedestrians with similar appearances.
	
	Moreover, we visualize the learned scores by DAM in Fig.~\ref{fig_dam}.
	As can be seen, the visible keypoints are more important than occluded ones; scores of visible frames are higher than those of occluded ones.
	Moreover, we found that the time-attention could learn more discriminative scores compared with the node-attention.

	\section{Conclusions}
	\label{conclusion}
	In this paper, we proposed to learn the pose feature to remedy the shortcomings of the appearance feature for the video-based ReID.
	To this end, we implemented a two-branch architecture to separately learn the pose feature and appearance feature.
	To learn the pose feature, we constructed a temporal pose graph via an off-the-shelf pose detector, whose nodes and edges denote the pose keypoints and skeleton connections, respectively.
	We then proposed a RGCN model to learn the node embeddings of the temporal graph, which could alleviate the influence from local node occlusion via the global information propagation mechanism.
	Finally, we employed the self-attention mechanism to obtain the temporal graph representation, where the node-attention and time-attention are leveraged to evaluate the importance of each node and each frame, respectively.
	We tested our method on the basis of multiple appearance models, and the experimental results demonstrated the superiority of the learned pose feature.

	\bibliographystyle{IEEEtran}
	\bibliography{IEEEabrv,mybibfile}

\end{document}